\documentclass[]{interact}
\usepackage{cite}
\usepackage{amsmath,amssymb,amsfonts}
\usepackage{graphicx}
\usepackage{subfigure}
\usepackage{textcomp}
\usepackage{stfloats}
\usepackage[noend]{algpseudocode}
\usepackage{algorithm,algorithmicx}
\usepackage{epstopdf}
\usepackage[caption=false]{subfig}

\theoremstyle{plain}

\theoremstyle{definition}

\theoremstyle{Remark}

\begin{document}

\articletype{ARTICLE TEMPLATE}

\title{Line Marching Algorithm For Planar Kinematic Swarm Robots: A Dynamic Leader-Follower Approach}

\author{
\name{He Cai\textsuperscript{a}, Shuping Guo\textsuperscript{a}, Yuheng He\textsuperscript{a}, Jieyi Yan\textsuperscript{a}, Yingnan Zhen\textsuperscript{a}, Huanli Gao\textsuperscript{a}, and Xiangyang Li\textsuperscript{a}\thanks{CONTACT Xiangyang Li Email: xyangli@scut.edu.cn}}
\affil{\textsuperscript{a}School of Automation Science and Engineering, South China University of Technology, Guangzhou 510641,  China.}
}

\maketitle

\begin{abstract}
Most of the existing formation algorithms for multiagent systems are fully label-specified, i.e., the desired
position for each agent in the formation is uniquely determined by its label, which would inevitably make the formation algorithms vulnerable to agent failures.
To address this issue, in this paper, we propose
a dynamic leader-follower approach to solving
the line marching problem for a swarm of planar kinematic robots.
In contrast to the existing results, the desired positions for the robots in the line are not fully label-specified, but determined in a dynamic way according to the current state of the robot swarm.
By constantly forming a chain of leader-follower pairs, exact formation can be
achieved by pairwise leader-following tracking.
Since the order of the chain of leader-follower pairs is constantly updated, the proposed algorithm shows strong robustness against robot failures.
Comprehensive numerical results are provided to evaluate the performance of the
proposed algorithm.
\end{abstract}


\section{Introduction}

Though each individual in a swarm is always subject to limited
sensing, computing, decision and execution capability,
the swarm as a whole exhibits marvelous performance.
For example, by working in groups, the ant colony can move food hundreds of times heavier than their weights,
thereby improving the foraging efficiency of the population \cite{mmt06}.
Another typical example is the formation flying of geese,
which helps to reduce energy expenditure and thus increase migration distance \cite{crp15}.
Inspired by the biological behaviors of social animals, formation control has received extensive attention
which gives rise to a significant amount of research results revealing the underlying mechanism of the biological behaviors.
In addition, formation control is also appealing from the perspective of engineering applications,
such as cooperative searching, exploration, escort and
commercial performance\cite{xwd16,fbk1,fbk2}.

Roughly speaking, there are two main classes of control approaches to solving the formation control problem. The first class is the
virtual structure  approach \cite{mk97,mm01,dpj07,lj12,fmhis22,cszl19,zb20},
which treats formation as a single entity, and the formation is maintained by keeping the
relative position between a virtual leader and each agent. Early works establishing the
virtual structure  approach can be found in \cite{mk97,mm01}.
Since after, the virtual structure  approach has been advanced in several directions,
 such as dealing with system nonlinearity \cite{dpj07}, integration with learning based approach \cite{lj12}, and
 adaptation to unreliable communication network \cite{fmhis22}, just to name a few. Recently,
a simulated annealing algorithm was proposed by \cite{cszl19} featuring optimal multiphase homing tracking
trajectory and a novel formation guidance strategy in order to reduce the impact caused by the unpredictable environment.
\cite{zb20} proposed a formation control strategy for UAVs, where the formation control problem was
decomposed into the smooth trajectory generation problem and the trajectory tracking problem.
With the help of the super-twisting method, the airflow influences between UAVs can be compensated effectively
when performing formation flying missions.
A salient advantage of the
virtual structure  approach lies in that it makes it easy to define, maintain and keep formation.
However, since the definition of formation is label-specified, once an agent fails, the formation will be incomplete unless further re-organization is made. As a result, the virtual structure  approach
is not robust or flexible enough in the face of agent failures.

The second class is the leader-follower approach \cite{dok01,arv02,lfd08,csyc10,ag18,hc19,cdl19,wsyw20,zrtc21}, which
decouples the formation control problem into a set of leader-following tracking problem. By specifying a
chain of leader-follower pair, formation can be achieved by the regulation of
relative position and direction of each leader-follower pair.
Seminal works for the leader-follower approach can be found in \cite{dok01,arv02},
which lay the foundation for various important extensions. For example,
\cite{lfd08} considered input constraint; \cite{csyc10} adopted a receding-horizon scheme
in the expect of fast exponential convergence; \cite{ag18} considered the fixed-time convergence issue;
\cite{hc19} took into account dynamic optimization; \cite{cdl19} studied distance-based formation.
Recently, in the absence of global position and direction measurement,
\cite{wsyw20}  proposed a flexible formation approach to track multiple mobile robots
based on estimated relative position. Likely, by  utilizing relative velocity and bearing measurements,
a distributed control approach was synthesized by \cite{zrtc21} which guaranteed a desired geometric pattern
for the agents under persistent exciting condition.
Due to the simple design philosophy, the leader-follower approach is easy to understand and implement in practice. While,
since the order of the chain of leader-follower pairs in the existing results are fixed, once an agent fails, the
other agents subsequential to the failed agent in the chain will be greatly affected, which makes the
traditional leader-follower approach vulnerable to agent failures.

As mentioned above, it is difficult for both the virtual structure approach and the traditional leader-follower approach
to deal with agent failures. The reason behind this is these approaches are fully label-specified, i.e., the desired
position for each agent in the formation is uniquely determined by its label.
To address this issue, in this paper, we further propose
a dynamic leader-follower approach to solving
the line marching problem for a swarm of planar kinematic robots.
In contrast to the existing results, the desired positions for the robots
in the line are not fully
label-specified, but determined in a dynamic way according to the current state of the robot swarm.
By constantly forming a chain of leader-follower pairs, exact formation can be
achieved by pairwise leader-following tracking.
Since the order of the chain of leader-follower pairs
is constantly updated, the proposed
algorithm shows strong robustness against robot failures.
In particular, once a robot fails, the rest
robots will re-form a new chain of leader-follower pairs and form a new marching line.
The failed robot will be considered as an
obstacle by other robots. After the failed robot gets back to normal, it will rejoin the swarm
and the whole robot swarm will re-form, again, into a new line and keep marching forward.

The following notation are adopted in this paper.
Let $\mathbb{R}$ and $\mathbb{N}$ denote the sets of real numbers and positive integers, respectively.
Given $x\in \mathbb{N}$,
\begin{equation*}
  \underline{x}\triangleq\{y|y\in \mathbb{N}, y\leq x\}.
\end{equation*}
Given angle $\theta$, the rotation matrix is defined as
\begin{equation*}
  \mathbf{R}(\theta)=\left(
                      \begin{array}{cc}
                        \cos(\theta) & -\sin(\theta) \\
                        \sin(\theta) & \cos(\theta) \\
                      \end{array}
                    \right).
\end{equation*}
 Given $x,y\in \mathbb{R}^2$, the inner product of $x$ and $y$
is denoted by $\langle x, y\rangle$.
The boolean relation ``or'' is denoted by $\vee$.
Given a nonzero vector $x\in \mathbb{R}^2$, let
$\gamma(x)=x/||x||$ denote the unit vector which has the same direction as $x$.

\section{System and task description}\label{secnot}

In this paper, we consider a swarm of $N$ planar kinematic
robots moving in an unbounded region subject to $M$ immobile or mobile
obstacles. Throughout this paper, it is assumed that
the positions of all the robots and obstacles are expressed in
a common coordinate system.

For $i\in\underline{N}$, the motion of the $i$th robot is described by the following equation:
\begin{equation}\label{robotdyn}
  \dot{p}_i(t)=v_i(t)
\end{equation}
where $p_i(t),v_i(t)\in \mathbb{R}^2$ denote the position and velocity vector of the $i$th robot, respectively.

Let $\delta_i>0$ denote the minimal safety distance for the $i$th robot.

For $i\in\underline{M}$, the motion of the $i$th obstacle is described by the following equation:
\begin{equation}\label{obdyn}
  \dot{q}_i(t)=u_i(t)
\end{equation}
where $q_i(t),u_i(t)\in \mathbb{R}^2$ denote the position and velocity vector of the $i$th obstacle, respectively. For immobile obstacles, $u_i(t)=0$ for all $t\geq 0$.

Let $\nu_i>0$ denote the minimal safety distance for the $i$th obstacle.

\begin{figure}[!htb]
\begin{center}
\scalebox{0.6}{\includegraphics[viewport=110 290 520 700]{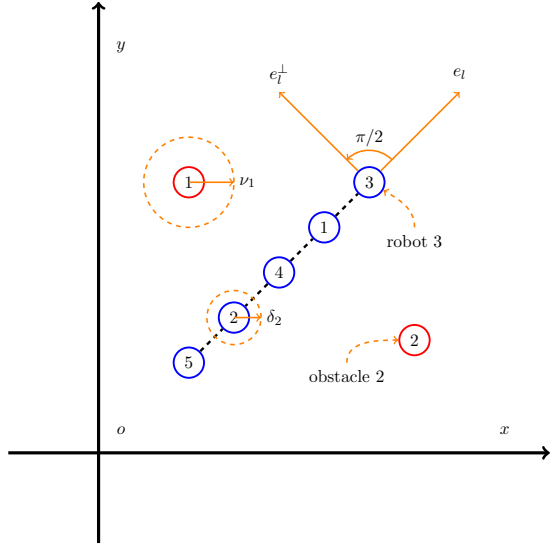}}
\caption{An example of line marching with 5 robots and 2 obstacles.}\label{line}
\end{center}

\end{figure}

Roughly speaking, as illustrated by Fig. \ref{line},
the task for the robots is to march in a line while
keep collision avoidance for both obstacles and other robots.
Let $v_l\in \mathbb{R}^2$ denote the nonzero marching velocity, and
$e_l=\gamma(v_l)$ denote the marching direction.
Then, let $e_l^{\perp}=\mathbf{R}(\pi/2)e_l$ denote the orientation vector
perpendicular to $e_l$ by rotating $e_l$ anticlockwise by $90$ degree.
Mathematically, the task considered in this paper can be described as follows:
\begin{itemize}
  \item collision avoidance
  \begin{enumerate}
    \item for all $t\geq 0$,
  for all $i,j\in\underline{N}$, $i\neq j$,
  \begin{equation}\label{taskca1}
    ||p_i(t)-p_j(t)||\geq \delta_i+\delta_j.
  \end{equation}
    \item for all $t\geq 0$, for all $i\in\underline{N}$, $j\in\underline{M}$,
  \begin{equation}\label{taskca2}
    ||p_i(t)-q_j(t)||\geq \delta_i+\nu_j.
  \end{equation}
  \end{enumerate}
  \item line marching
  \begin{enumerate}
    \item for all $i\in\underline{N}$,
    \begin{equation}\label{tacklm1}
      \lim_{t\rightarrow\infty}(v_i(t)-v_l)=0.
    \end{equation}
    \item for all $i,j\in\underline{N}$, $i\neq j$,
    \begin{equation}\label{tacklm2}
      \lim_{t\rightarrow\infty}\langle p_i(t)-p_j(t),e_l^{\perp}\rangle=0.
    \end{equation}
    \item there exists an order $(i_1,i_2,\dots,i_N)$ where $i_k\in \underline{N}$ such that
    for $k\in \underline{N-1}$,
    \begin{equation}\label{tacklm3}
      \lim_{t\rightarrow\infty}\langle p_{i_{k+1}}(t)-p_{i_k}(t),e_l\rangle=\rho
    \end{equation}
    where $\rho>0$ denotes the desired inter-robot distance.
  \end{enumerate}
\end{itemize}
To avoid the possible conflict between collision avoidance and
line marching, $\rho$ should be chosen such that for all $i,j\in \underline{N}$,
$i\neq j$,
\begin{equation*}
  \rho>\delta_i+\delta_j.
\end{equation*}

  Equation \eqref{tacklm3} indicates that the order of the marching line is immaterial as long as the relative
  position of the robots in the line can be maintained, which makes it possible that, when some robots
  fail, the rest robots can re-form a new marching line.

\section{System modeling and problem formulation}\label{secsmpf}

In this paper, we propose a novel dynamic leader-follower approach for the task of robots line marching.
The basic idea is to form a chain of leader-follower pairs in a self-organized way in the sense
that the order of the chain shall not be uniquely specified by the label of the robots, but rather
determined by the current state of the robot swarm.

\subsection{Preliminaries}\label{secmain}

To begin with, we will introduce some indices which will be used subsequently.
\begin{enumerate}
  \item to indicate whether robot $i$ is the head of the line or not, the following
  head index is needed:
  \begin{equation}\label{}
  \Lambda_i=\left\{
  \begin{array}{ll}
    1, & \hbox{robot $i$ is the head of the line;} \\
    0, & \hbox{otherwise.}
  \end{array}
  \right.
\end{equation}
  \item to indicate whether robot $i$ has a follower or not, the following
  follower index is needed:
  \begin{equation}\label{}
  \Delta_i=\left\{
  \begin{array}{ll}
    1, & \hbox{robot $i$ has a follower;} \\
    0, & \hbox{otherwise.}
  \end{array}
  \right.
\end{equation}
  \item to indicate whether robot $i$ fails or not, the following
  failure index is needed:
  \begin{equation}\label{}
  \Gamma_i=\left\{
  \begin{array}{ll}
    0, & \hbox{robot $i$ fails;} \\
    1, & \hbox{otherwise.}
  \end{array}
  \right.
\end{equation}
 \item to indicate whether robot $i$ has a leader or not, the following
  leader index is needed:
  \begin{equation}\label{}
  \Phi_i=\left\{
  \begin{array}{ll}
    1, & \hbox{robot $i$ has a leader;} \\
    0, & \hbox{otherwise.}
  \end{array}
  \right.
\end{equation}
\end{enumerate}

To achieve simultaneous line marching and collision avoidance,
$v_i(t)$ of equation \eqref{robotdyn} consists of two parts:
\begin{equation}\label{}
  v_i(t)=v_{i,lm}(t)+v_{i,ca}(t)
\end{equation}
where $v_{i,lm}(t),v_{i,ca}(t)\in \mathbb{R}^2$ aim at line marching and
collision avoidance, respectively.

To avoid collision, for robot $i$, we let
\begin{equation}\label{}
\begin{aligned}
  &\bar{v}_{i,ca}(t)\\
  =&\sum_{j=1}^N\gamma(p_i(t)-p_j(t))\zeta\left(||p_i(t)-p_j(t)||,\delta_i,\delta_j,\kappa_1,\kappa_2\right)\\
  +&\sum_{j=1}^M\gamma(p_i(t)-q_j(t))\zeta\left(||p_i(t)-q_j(t)||,\delta_i,\nu_j,\kappa_1,\kappa_2\right)
\end{aligned}
\end{equation}
where $\kappa_1>1$, $\kappa_2>0$, and
\begin{equation}\label{}
\begin{aligned}
  &\zeta(x,a,b,\kappa_1,\kappa_2)\\
=&\left\{
                                   \begin{array}{ll}
                                     \left(\frac{\kappa_2}{x-(a+b)}
  -\frac{\kappa_2}{(\kappa_1-1)(a+b)}\right), & a+b<x\leq \kappa_1(a+b) \\
                                     0, & x>\kappa_1(a+b)
                                   \end{array}
                                 \right.
\end{aligned}
\end{equation}

To rule out the case that
the robot is stuck at some place when $v_{i,lm}(t)+\bar{v}_{i,ca}(t)=0$, we let
\begin{equation}\label{vica}
  v_{i,ca}(t)=\mathbf{R}(\theta_i(t))\bar{v}_{i,ca}(t)
\end{equation}
where
\begin{equation}
  \theta_i(t)=a_i\cdot\frac{\pi}{180}\cdot\sin \omega_i t
\end{equation}
where $a_i,\omega_i>0$ are design parameters.
By choosing $a_i$ sufficiently small, the orientation of
$\bar{v}_{i,ca}(t)$ shall constantly swing at a small amplitude, which
would almost surely avoid the case of $v_{i,lm}(t)+\bar{v}_{i,ca}(t)=0$.

For $j\neq i$, we let
\begin{equation}\label{vilmj}
\begin{aligned}
   v_{i,lm}^j(t)&=v_l+\alpha(\langle p_j(t)-p_i(t),e_l\rangle-\rho) e_l\\
  &+\beta \langle p_j(t)-p_i(t),e_l^{\perp}\rangle e_l^{\perp}
\end{aligned}
\end{equation}
where $\alpha,\beta>0$ are control gains.

\begin{algorithm}\label{ag1}
\caption{Line Marching Algorithm} 
\hspace*{0.02in} {\bf Input:} 
$p_1(t),\dots,p_N(t)$.\\
\hspace*{0.02in} {\bf Output:} 
$v_1(t),\dots,v_N(t)$.
\begin{algorithmic}[1]
\For{$i=1$, $i\leq N$, $i=i+1$}
\State set $\Delta_i=0$, $\Phi_i=0$, $\Lambda_i=1$.
\State set $v_i(t)=0$.
\EndFor

\For{$i=1$, $i\leq N$, $i=i+1$}

\If{$\Gamma_i=1$}

\For{$j=1$, $j\leq N$, $j\neq i$, $j=j+1$ \& $\Phi_i=0$ }
\If {$\langle p_j(t)-p_i(t),e_l\rangle>0$ \& $\Gamma_j=1$}
\State set $\Lambda_i=0$.
\If {$\Delta_j=0$}
\State $v_i(t)=v_{i,lm}^j(t)+v_{i,ca}(t)$,
\State set $\Delta_j=1$, $\Phi_i=1$.
\EndIf
\EndIf
\EndFor

\If {$\Lambda_i=1$}

\If {$i=1$}
\State $v_i(t)=v_l+v_{i,ca}(t)$.
\ElsIf {$\Lambda_1\vee\cdots\vee\Lambda_{i-1}=0$}
\State $v_i(t)=v_l+v_{i,ca}(t)$.
\EndIf

\EndIf

\EndIf

\EndFor
\State \Return result
\end{algorithmic}
\end{algorithm}

\subsection{Algorithm design}

Now, we are ready to present the linear marching algorithm by Algorithm 1.
The basic process is as follows.
Initially, all the robots consider themselves as the head of the line and
the velocity by default is set to be zero. After the initialization phase, the decision phase
starts.
Robot $i$ checks the relative position
between itself and other robots in the ascending order of the robot label.
If there is at least one robot ahead of it along the marching direction, then
robot $i$ changes the status of its head index. Moreover, if robot $i$ further finds a robot, say, robot $j$,
ahead of it along the marching direction, which, in the meantime, has not been tailed,
then robot $j$ will be taken as the leader for robot $i$, and the decision loop of robot $i$ ends.
If a robot is the foremost one of the line, it automatically becomes the head of the line. If there are multiple robots situate as the foremost heads of the ensemble, only the one with the smallest label will be selected as the leader of the line and others will stay still temporarily  to avoid the co-head situation. If a robot fails, it will also stay
where it is until it gets back to normal.

\subsection{Discussions}\label{seccd}

Algorithm 1 can be either implemented in a centralized way, that is, there exists
a central controller taking control of the whole robot swarm, or implemented in a decentralized way,
that is, each robot makes its own decision based on inter-robot communication.
In particular, each robot is endowed with its own time slice to receive information,
make decision and then send out updated information, and
all the robots share a common cyclic communication loop.
The reason behind the requirement on the sequential way of information gathering and broadcasting
lies in the demand for exact formation, i.e., the inter-agent
distance in the marching line should be strictly equal to the designed value.
To the best of the authors' knowledge, under the distributed communication mechanism,
it would be rather difficult, if possible, to simultaneously achieve exact formation and
autonomous formation reconfiguration when robots fail.

It is also worth mentioning that, in practice, the robots are always subject to
noises and external disturbances. As a result, the co-head situation will not happen.
Therefore, Algorithm 1 can be further
simplified to Algorithm 3, which will be shown later in Section \ref{sec.apptoros}.

The design of Algorithm 1 indicates that the priority of
collision avoidance is higher than line marching for the reason that $v_{i,lm}(t)$
is always bounded, while $\bar{v}_{i,ca}(t)$ would go unbounded
when the distance between robot and obstacle, or two robots,
approaches the safety limit.

\begin{figure}
\begin{center}
\scalebox{0.65}{\includegraphics[viewport=23 51 406 273, clip]{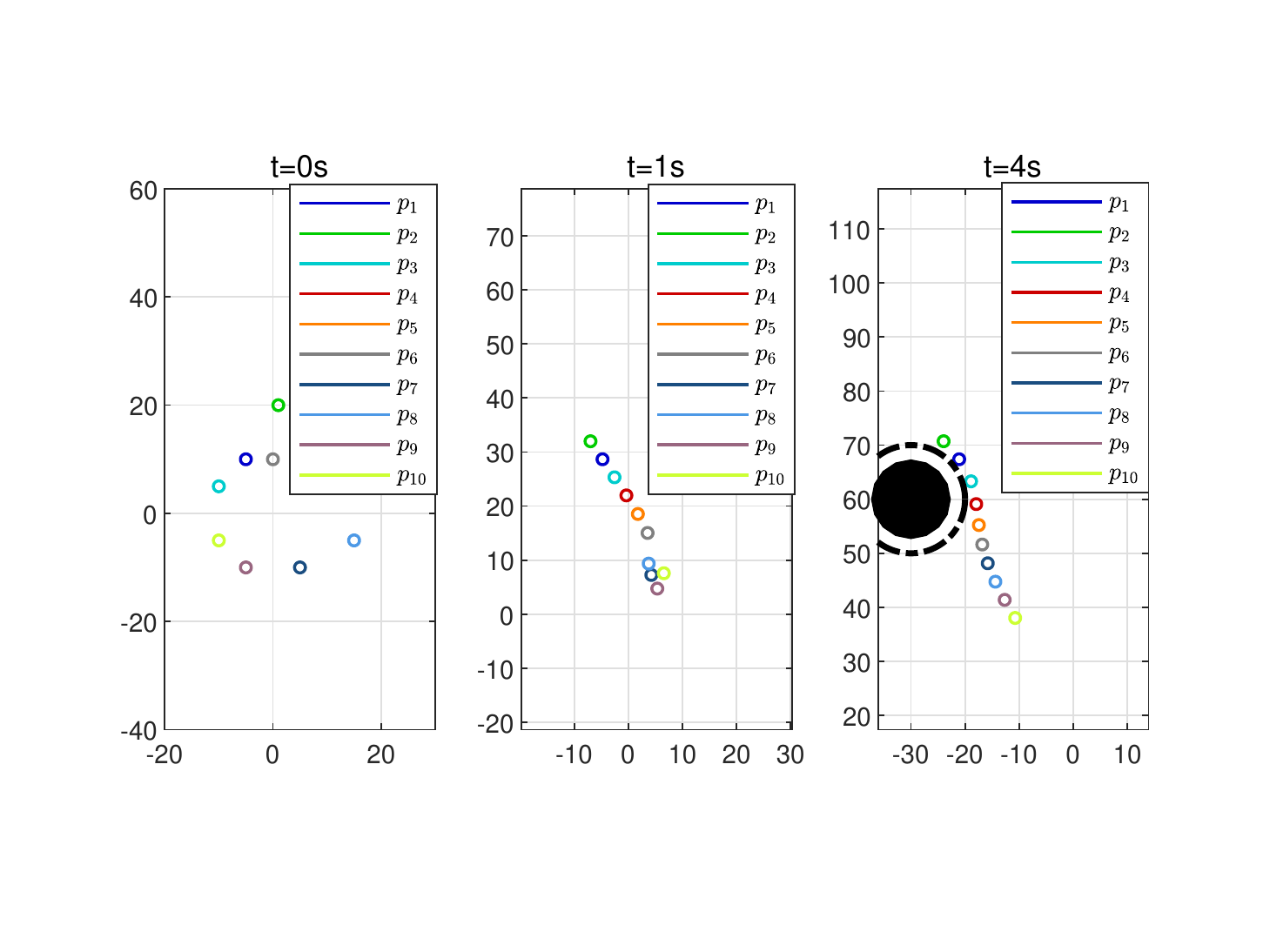}}
\scalebox{0.65}{\includegraphics[viewport=23 51 406 273, clip]{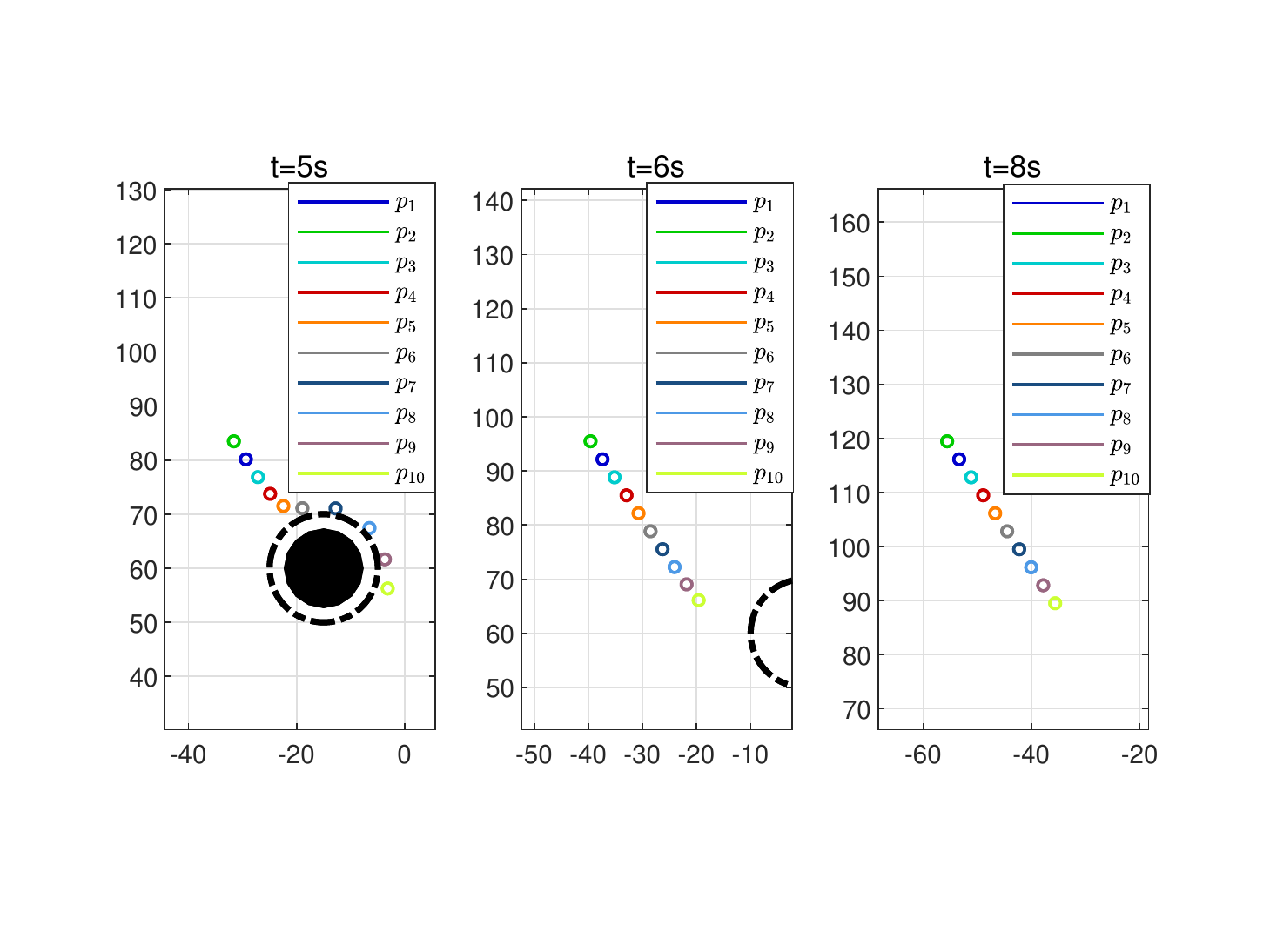}}
\caption{Robots' and mobile obstacle's positions in Case \ref{subsec.casea} at different time instants.}\label{casea1}
\end{center}

\end{figure}

\begin{figure*}
  \begin{center}

  \subfigure{
\begin{minipage}[t]{0.5\linewidth}
\centering
\scalebox{0.45}{\includegraphics{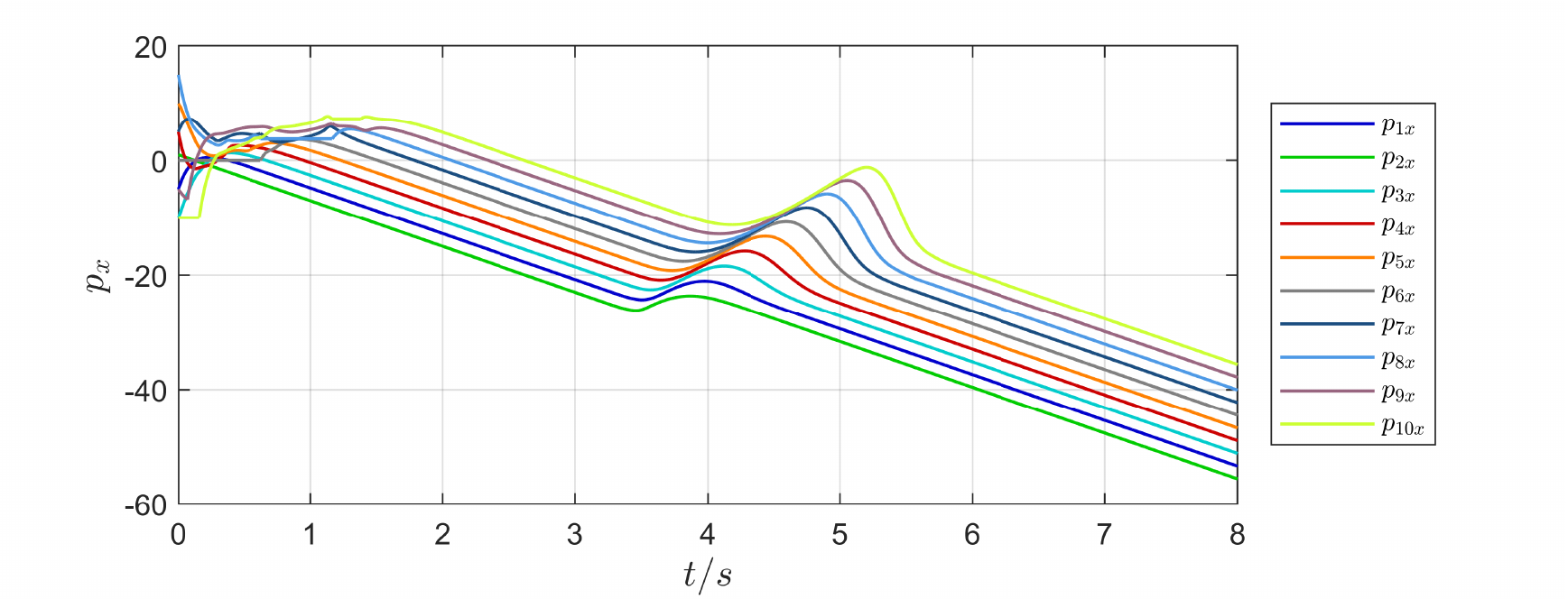}}
\end{minipage}%
}%
\subfigure{
\begin{minipage}[t]{0.5\linewidth}
\centering
\scalebox{0.45}{\includegraphics{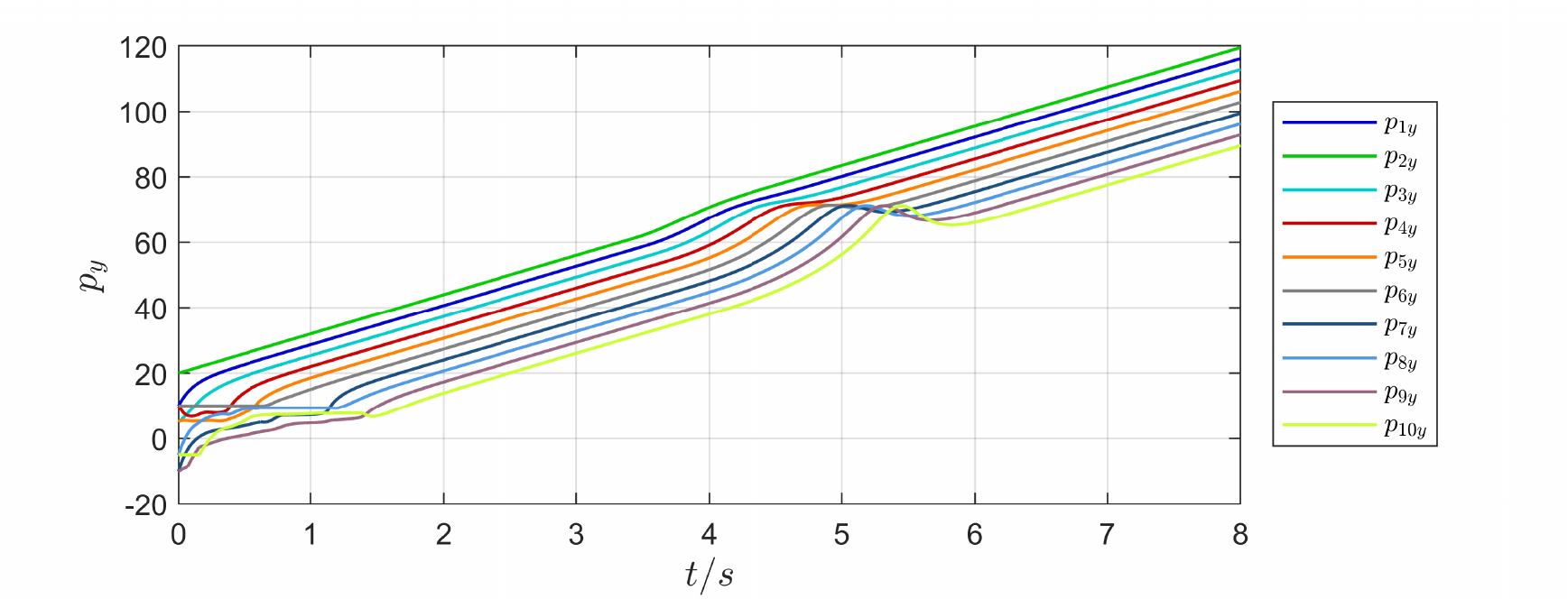}}
\end{minipage}%
}%

\end{center}
  \caption{Time profiles of robots' positions in Case \ref{subsec.casea}.}\label{casea21}
  \vspace{-0.5cm}
\end{figure*}

\begin{figure*}
  \begin{center}

  \subfigure{
\begin{minipage}[t]{0.5\linewidth}
\centering
\scalebox{0.45}{\includegraphics{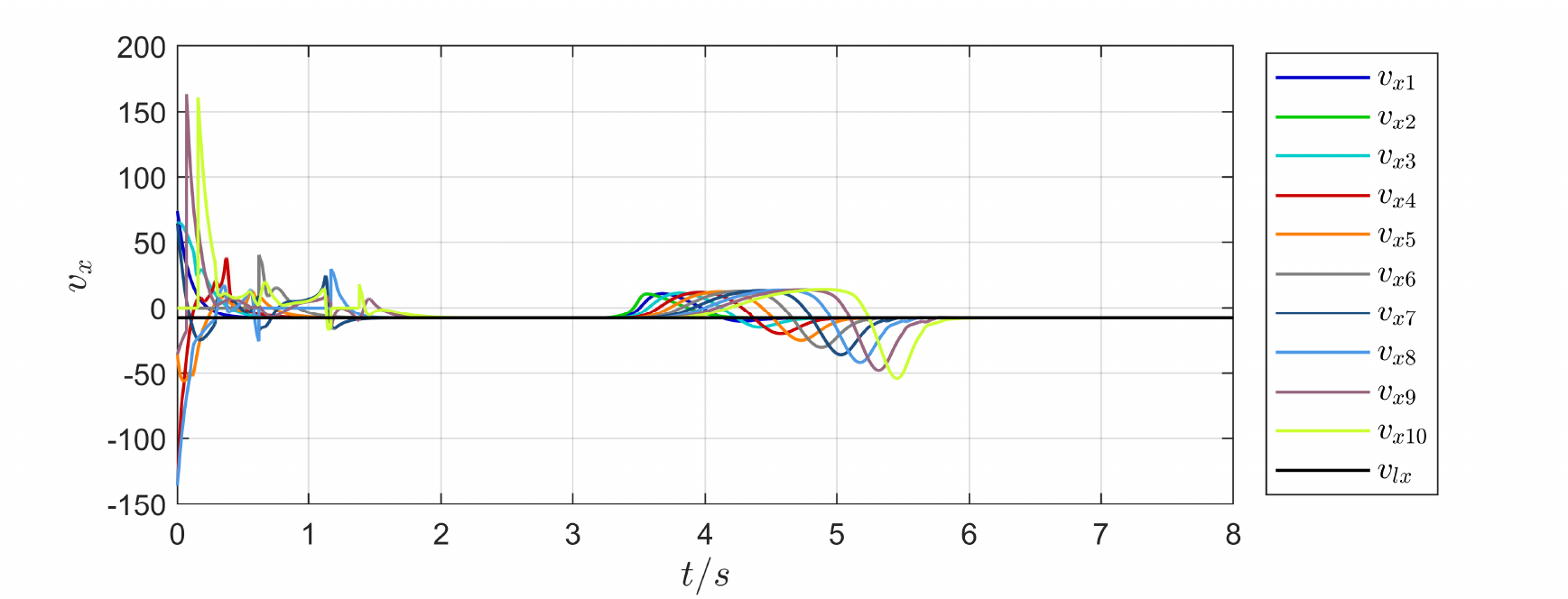}}
\end{minipage}%
}%
\subfigure{
\begin{minipage}[t]{0.5\linewidth}
\centering
\scalebox{0.45}{\includegraphics{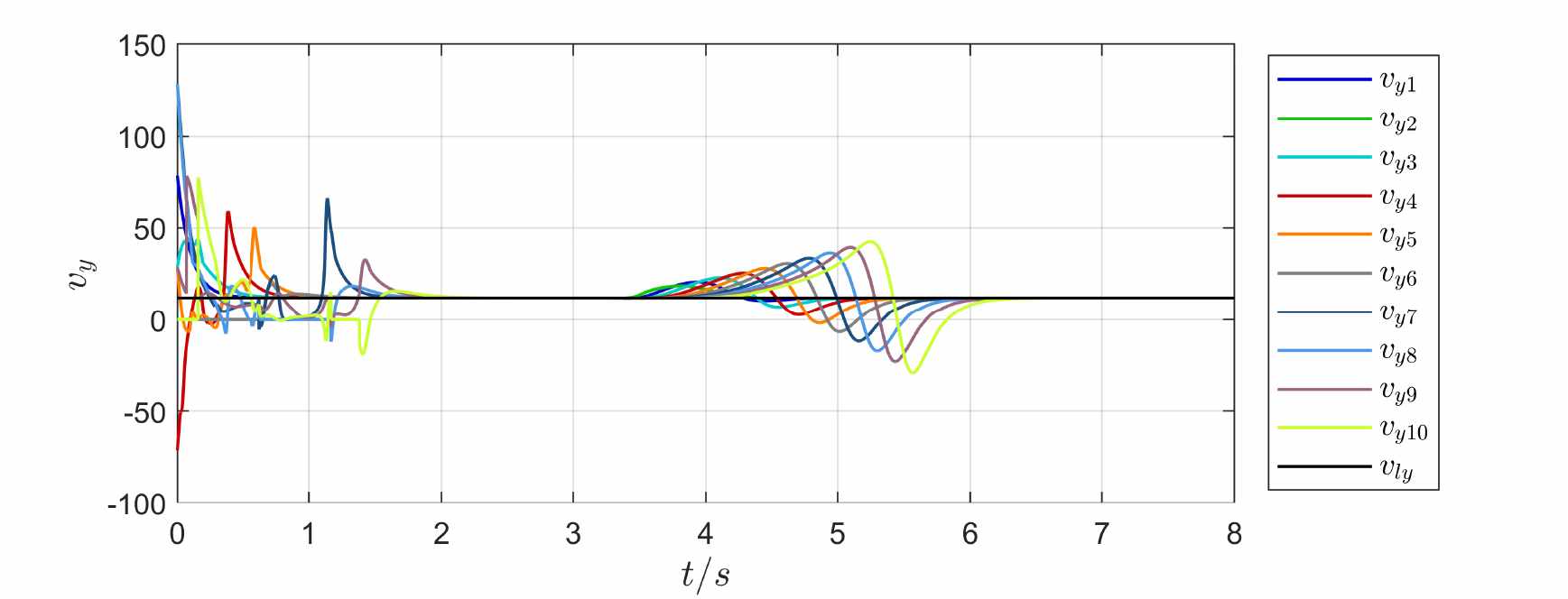}}
\end{minipage}%
}%

\end{center}
  \caption{Time profiles of robots' velocities in Case \ref{subsec.casea}.}\label{casea22}
  \vspace{-0.5cm}
\end{figure*}

\begin{figure*}
  \begin{center}

  \subfigure{
\begin{minipage}[t]{0.5\linewidth}
\centering
\scalebox{0.45}{\includegraphics{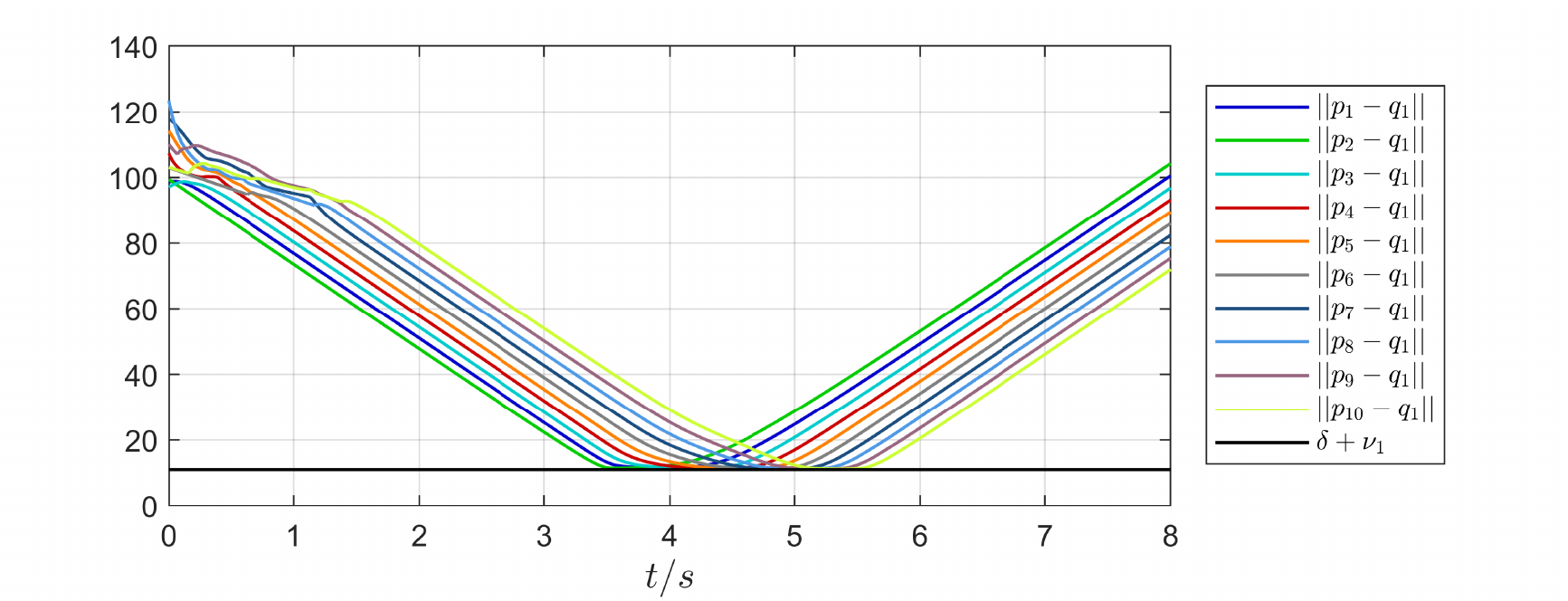}}
\end{minipage}%
}%
\subfigure{
\begin{minipage}[t]{0.5\linewidth}
\centering
\scalebox{0.45}{\includegraphics{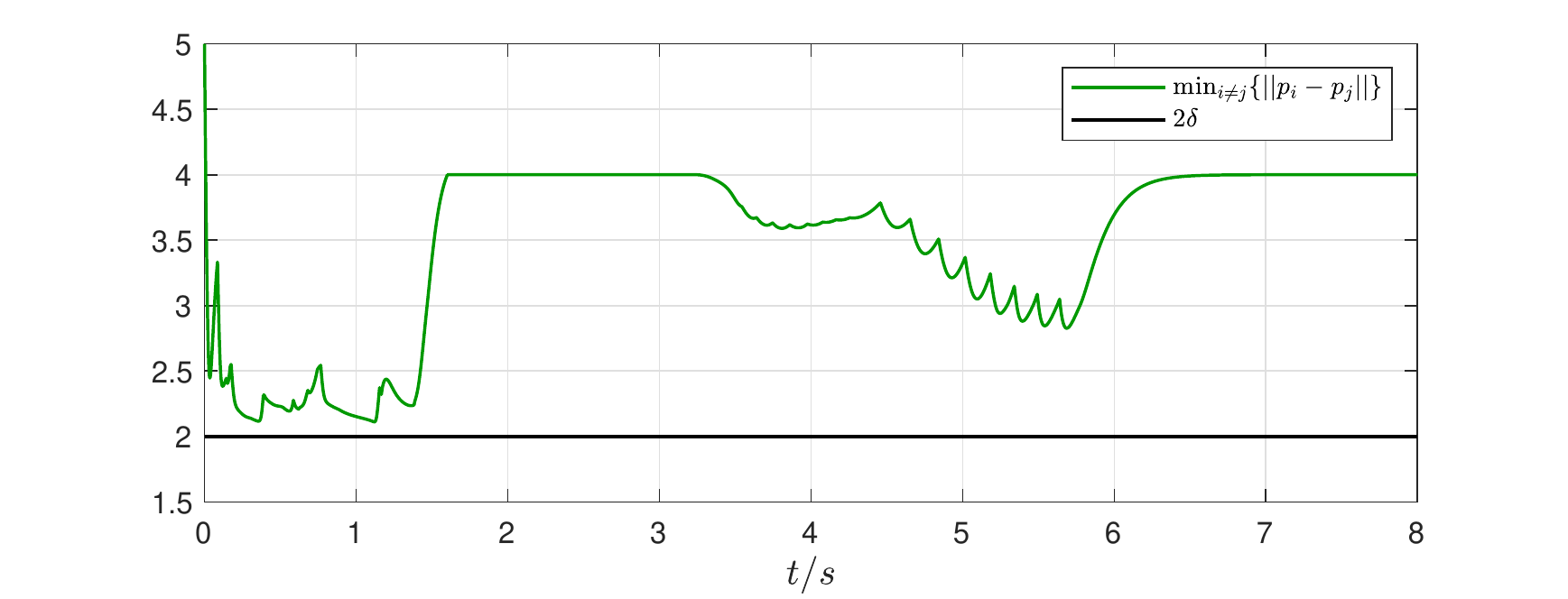}}
\end{minipage}%
}%
\end{center}

  \caption{Time profiles of the relative distance between each robot
and the mobile obstacle, and the minimal relative distance between two robots in Case \ref{subsec.casea}.}\label{casea23}
\vspace{-0.5cm}
\end{figure*}

\begin{figure*}
  \begin{center}

  \subfigure{
\begin{minipage}[t]{0.5\linewidth}
\centering
\scalebox{0.45}{\includegraphics{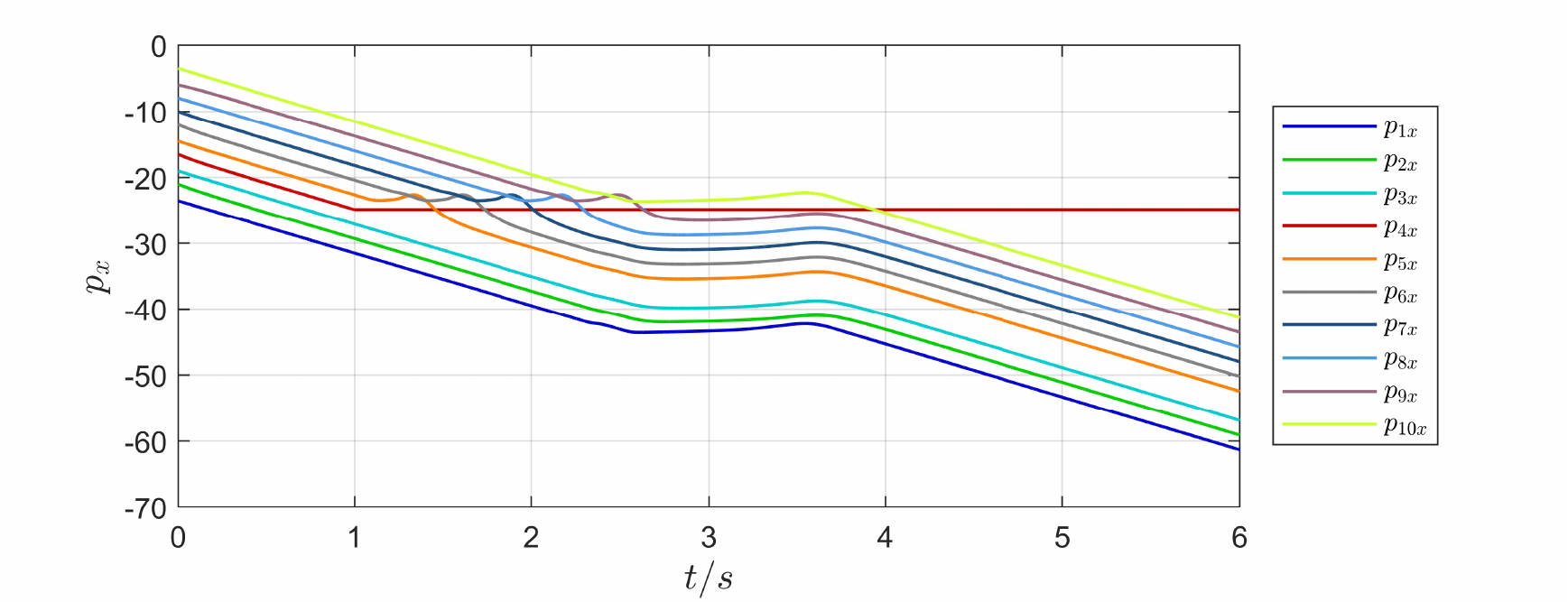}}
\end{minipage}%
}%
\subfigure{
\begin{minipage}[t]{0.5\linewidth}
\centering
\scalebox{0.45}{\includegraphics{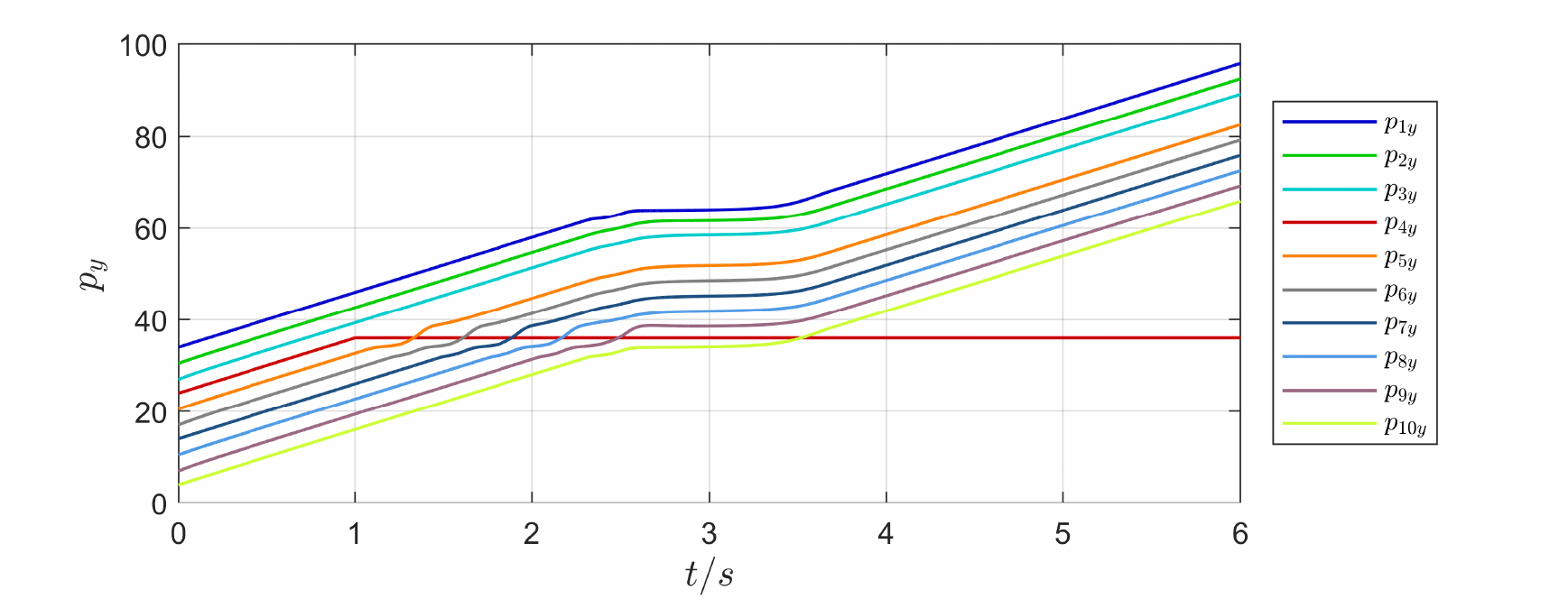}}
\end{minipage}%
}%

\end{center}
  \caption{Time profiles of robots' positions using the virtual structure approach when robot $4$ fails
  at $t=1$s in Case \ref{subsec.caseb}.}\label{caseb1}
  \vspace{-0.5cm}
\end{figure*}

\begin{figure*}
  \begin{center}

  \subfigure{
\begin{minipage}[t]{0.5\linewidth}
\centering
\scalebox{0.45}{\includegraphics{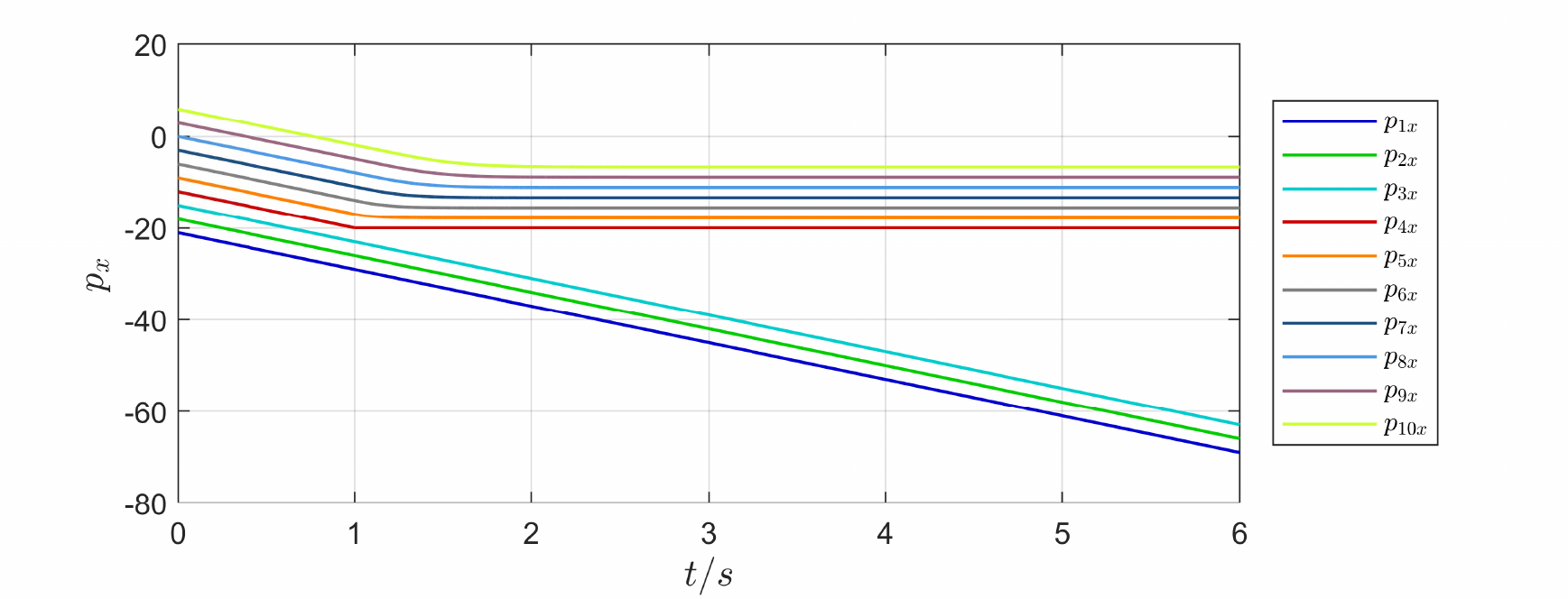}}
\end{minipage}%
}%
\subfigure{
\begin{minipage}[t]{0.5\linewidth}
\centering
\scalebox{0.45}{\includegraphics{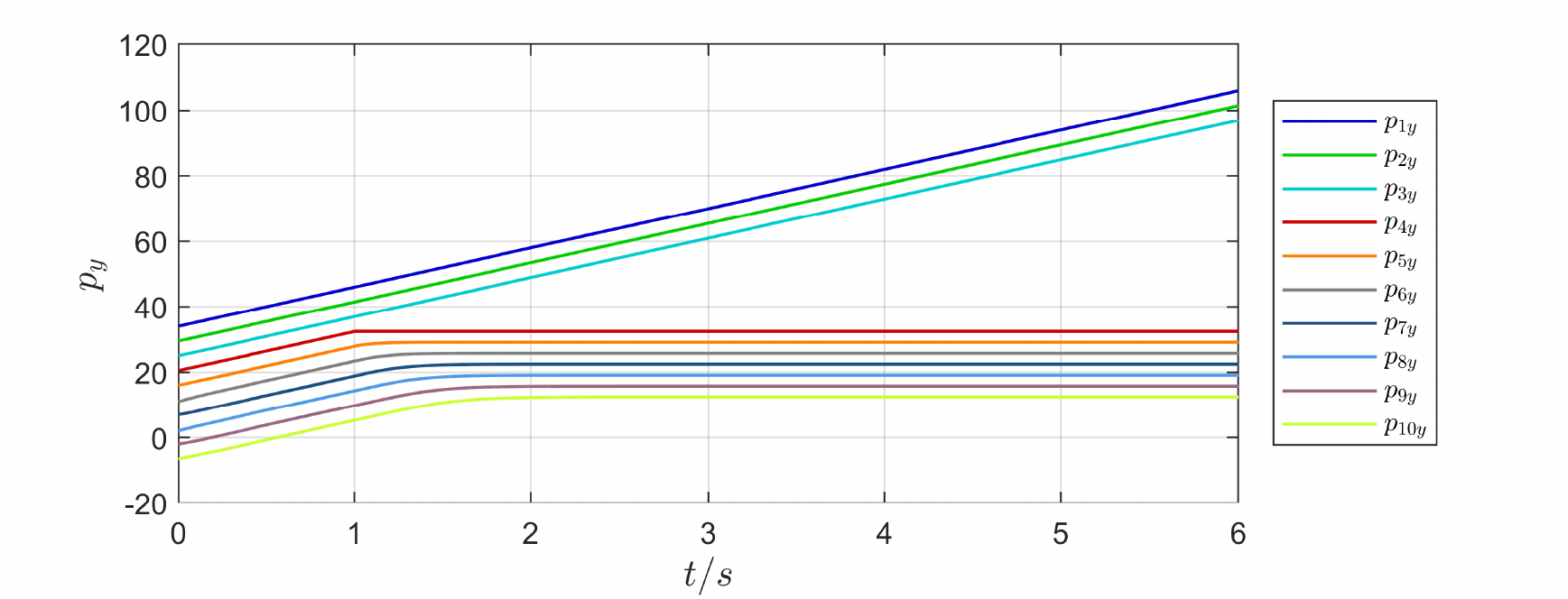}}
\end{minipage}%
}%

\end{center}
  \caption{Time profiles of robots' positions using the traditional leader-follower approach when robot $4$ fails
  at $t=1$s in Case \ref{subsec.caseb}.}\label{caseb2}
  \vspace{-0.5cm}
\end{figure*}

\begin{figure*}
  \begin{center}

  \subfigure{
\begin{minipage}[t]{0.5\linewidth}
\centering
\scalebox{0.45}{\includegraphics{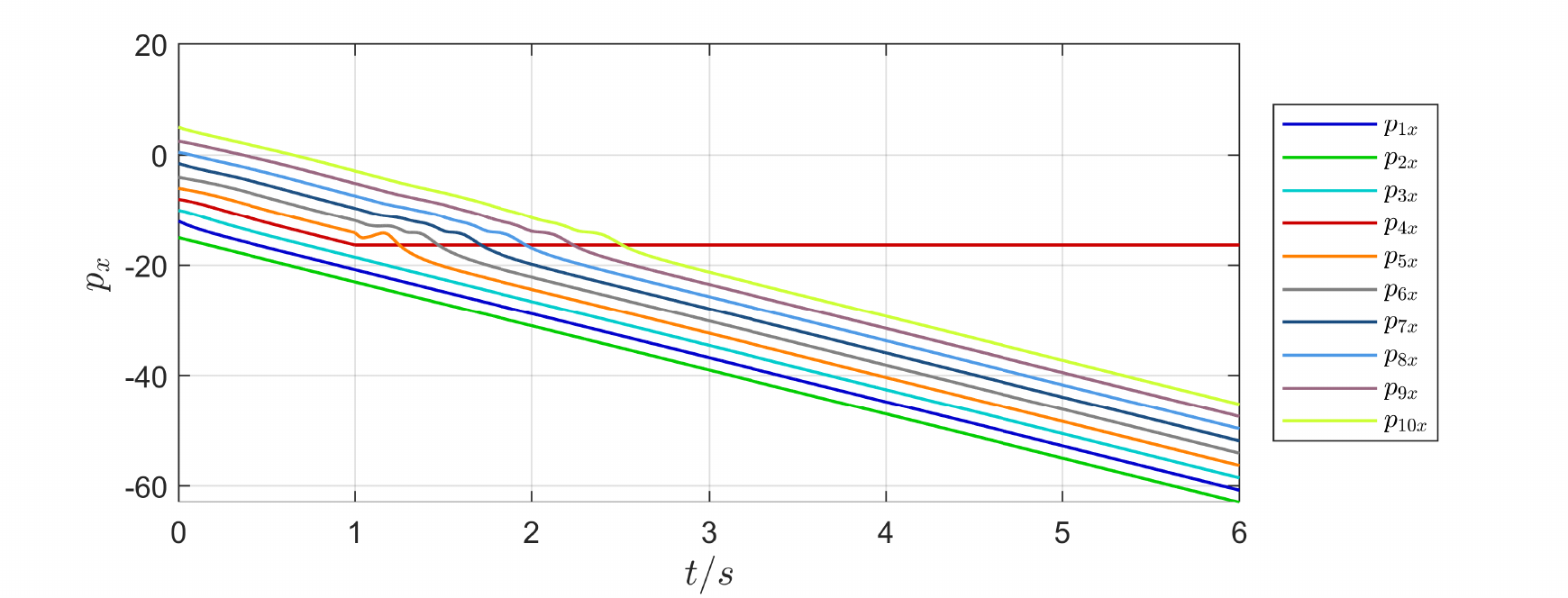}}
\end{minipage}%
}%
\subfigure{
\begin{minipage}[t]{0.5\linewidth}
\centering
\scalebox{0.45}{\includegraphics{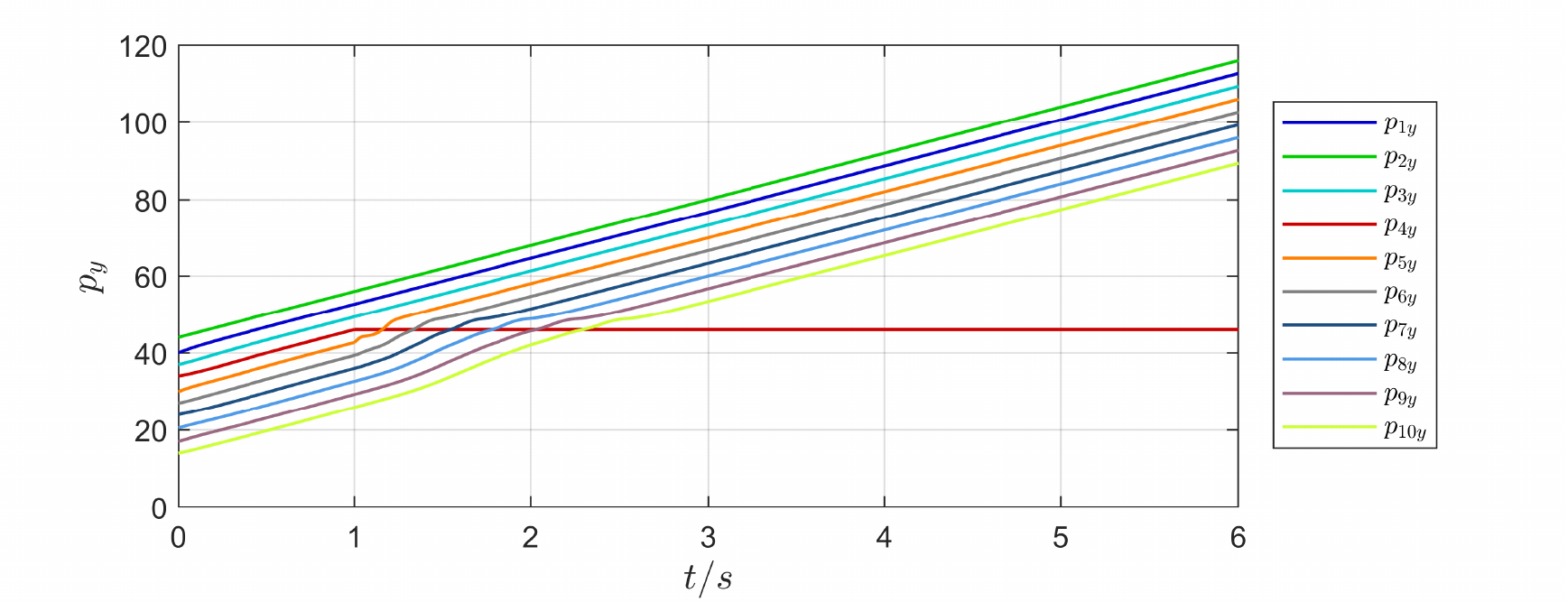}}
\end{minipage}%
}%

\end{center}
  \caption{Time profiles of robots' positions using the dynamic leader-follower approach when robot $4$ fails
  at $t=1$s in Case \ref{subsec.caseb}.}\label{caseb3}
\end{figure*}

\section{Numerical simulations}\label{sec.apptocon}
In this section, we consider a swarm of $10$ planar kinematic robots.
The parameters for the line marching are given by $v_l=(-8,12)^T\triangleq (v_{lx},v_{ly})^T$, $\rho=4$.
For simplicity, the minimal inter-robot safety distance is set to be $\delta_i\triangleq\delta=1$.
The control parameters are selected to be $\kappa_1=1.5$, $\kappa_2=10$,
$a_i=10$, $\omega_i=1$, $\alpha=10$, $\beta=10$.
In the following, we will consider two cases for numerical simulation.

\subsection{In the presence of a mobile obstacle}\label{subsec.casea}
In this case, we consider the line marching subject to a mobile obstacle with $\nu_1=10$ and $u_1(t)=(15,0)^T$.
The initial positions of the robots are give by
$p_1(0)=(-5,10)^T$, $p_2(0)=(1,20)^T$, $p_3(0)=(-10,5)^T$, $p_4(0)=(5,10)^T$, $p_5(0)=(10,5)^T$, $p_6(0)=(0,10)^T$, $p_7(0)=(5,-10)^T$, $p_8(0)=(15,-5)^T$, $p_9(0)=(-5,-10)^T$, $p_{10}(0)=(-10, -5)^T$. The initial position for the
obstacle is given by $q_1(0)=(-90,60)^T$.

Fig. \ref{casea1} shows the robots' and mobile obstacle's positions at different time instants.
Figs. \ref{casea21} and \ref{casea22} show the time profiles of robots' positions and velocities, respectively.
It can be seen that the line marching objective has been successfully achieved.
Fig. \ref{casea23} shows the relative distance between each robot
and the mobile obstacle, and the minimal relative distance between two robots. It can be
seen that all the lines are above the associated safety limits, indicating that
there is no collision during the whole process.

\subsection{In the presence of robot failure}\label{subsec.caseb}
In this case, we consider the scenario when some robot fails and show the
comparisons between the virtual structure approach, the traditional leader-follower
approach, and the dynamic leader-follower approach proposed in this paper.
For simplicity, initially, we assume the robots are already in line formation.
In particular, we let $p_1(0)=(-12,40)^T$, $p_2(0)=(-15,44)^T$, $p_3(0)=(-10,37)^T$, $p_4(0)=(-8,34)^T$, $p_5(0)=(-6,30)^T$, $p_6(0)=(-4,27)^T$, $p_7(0)=(-1.5,24)^T$, $p_8(0)=(0.5,20.5)^T$, $p_9(0)=(2.5,-17)^T$, $p_{10}(0)=(5, 14)^T$.
Suppose at $t=1$s, robot $4$ fails.

The system performance using the virtual structure approach is shown by Fig. \ref{caseb1}. Since the desired position
by the virtual structure approach is uniquely determined by the label of the robot, when robot $4$ fails, there will be a vacancy
in the formation which cannot be made up by other robots unless further formation reconfiguration
is invoked.

The system performance using the traditional leader-follower approach is shown by Fig. \ref{caseb2}. The order of the
chain of leader-follower pairs is simply set to be $(1,2)$, $(2,3)$, $(3,4)$, $(4,5)$, $(5,6)$, $(6,7)$, $(7,8)$,
$(8,9)$, $(9,10)$. It can be seen from the simulation results that, when robot $4$ fails, robots $5-10$ all get stuck since the
order of the chain of leader-follower pairs is fixed for the traditional leader-follower approach.

The system performance using the dynamic leader-follower approach proposed in this paper is shown by Fig. \ref{caseb3}.
It can be seen from the simulation results that, when robot $4$ fails, the rest robots has re-formed a new chain of
leader-follower pairs, bypassed robot $4$, and formed a new marching line, which shows strong robustness against robot
failures.

\section{Application to the discrete-time model}\label{sec.apptodis}

In this section, we apply the proposed line marching algorithm to
discrete-time model. In this case, equations \eqref{robotdyn} and \eqref{obdyn} become
\begin{equation}\label{}
  p_i(k+1)=p_i(k)+Tv_i(k)
\end{equation}
and
\begin{equation}\label{}
  q_i(k+1)=q_i(k)+Tu_i(k)
\end{equation}
respectively, where $T>0$ denotes the sampling time.
Note that the discrete-time versions of $v_{i,ca}(t)$ and
$v_{i,lm}^j(t)$ by \eqref{vica} and \eqref{vilmj} are straightforward and thus are omitted.
\begin{algorithm}[h]\label{ag2}
\caption{Line Marching Algorithm (Discrete-Time)} 
\hspace*{0.02in} {\bf Input:} 
$p_1(k),\dots,p_N(k)$.\\
\hspace*{0.02in} {\bf Output:} 
$v_1(k),\dots,v_N(k)$.
\begin{algorithmic}[1]
\For{$i=1$, $i\leq N$, $i=i+1$}
\State set $\Delta_i=0$, $\Phi_i=0$, $\Lambda_i=1$.
\State set $v_i(k)=0$.
\EndFor
\For{$i=1$, $i\leq N$, $i=i+1$}
\If{$\Gamma_i=1$}
\For{$j=1$, $j\leq N$, $j\neq i$, $j=j+1$ \& $\Phi_i=0$ }
\If {$\langle p_j(k)-p_i(k),e_l\rangle>0$ \& $\Gamma_j=1$}
\State set $\Lambda_i=0$.
\If {$\Delta_j=0$}
\State $v_i(k)=v_{i,lm}^j(k)+v_{i,ca}(k)$,
\State set $\Delta_j=1$, $\Phi_i=1$.
\EndIf
\EndIf
\EndFor
\If {$\Lambda_i=1$}
\If {$i=1$}
\State $v_i(k)=v_l+v_{i,ca}(k)$.
\ElsIf {$\Lambda_1\vee\cdots\vee\Lambda_{i-1}=0$}
\State $v_i(k)=v_l+v_{i,ca}(k)$.
\EndIf
\EndIf
\EndIf
\EndFor
\State \Return result
\end{algorithmic}
\end{algorithm}
The line marching algorithm for the discrete-time case is given by
Algorithm 2. In what follows, we use Python to obtain the simulation results.

In this section, we also consider a swarm of $10$ planar kinematic robots.
The parameters for the line marching are given by $v_l=(-8,12)^T$, $\rho=4$.
The sampling time is chosen to be $T=0.001$. The minimal inter-robot safety distance is set to be $\delta_i\triangleq\delta=1$.
The control parameters are selected to be $\kappa_1=1.5$, $\kappa_2=10$,
$a_i=10$, $\omega_i=1$, $\alpha=10$, $\beta=10$.

We consider the line formation subject to a mobile obstacle with $\nu_1=10$ and $u_1(t)=(15,0)^T$.
The initial positions of the robots are give by
$p_1(0)=(-5,10)^T$, $p_2(0)=(1,20)^T$, $p_3(0)=(-10,5)^T$, $p_4(0)=(5,10)^T$, $p_5(0)=(10,5)^T$, $p_6(0)=(0,10)^T$, $p_7(0)=(5,-10)^T$, $p_8(0)=(15,-5)^T$, $p_9(0)=(-5,-10)^T$, $p_{10}(0)=(-10, -5)^T$. The initial position for the
obstacle is given by $q_1(0)=(-80,60)^T$.

In the following simulation, we consider four phases:
\begin{enumerate}
  \item from $k=0$ to $k=3000$, the robots will form a line from the initial position;
  \item from $k=3000$ to $k=8000$, the robots will bypass the mobile obstacle and re-form a marching line;
  \item from $k=8000$ to $k=11000$, robot 4 fails;
  \item from $k=11000$ to $k=15000$, robot 4 gets back to normal and rejoins the robot swarm.
\end{enumerate}

\begin{figure*}
\begin{center}

\subfigure{
    \begin{minipage}[t]{0.48\linewidth}
    \centering
    \scalebox{0.4}{\includegraphics[viewport=60 250 550 620, clip]{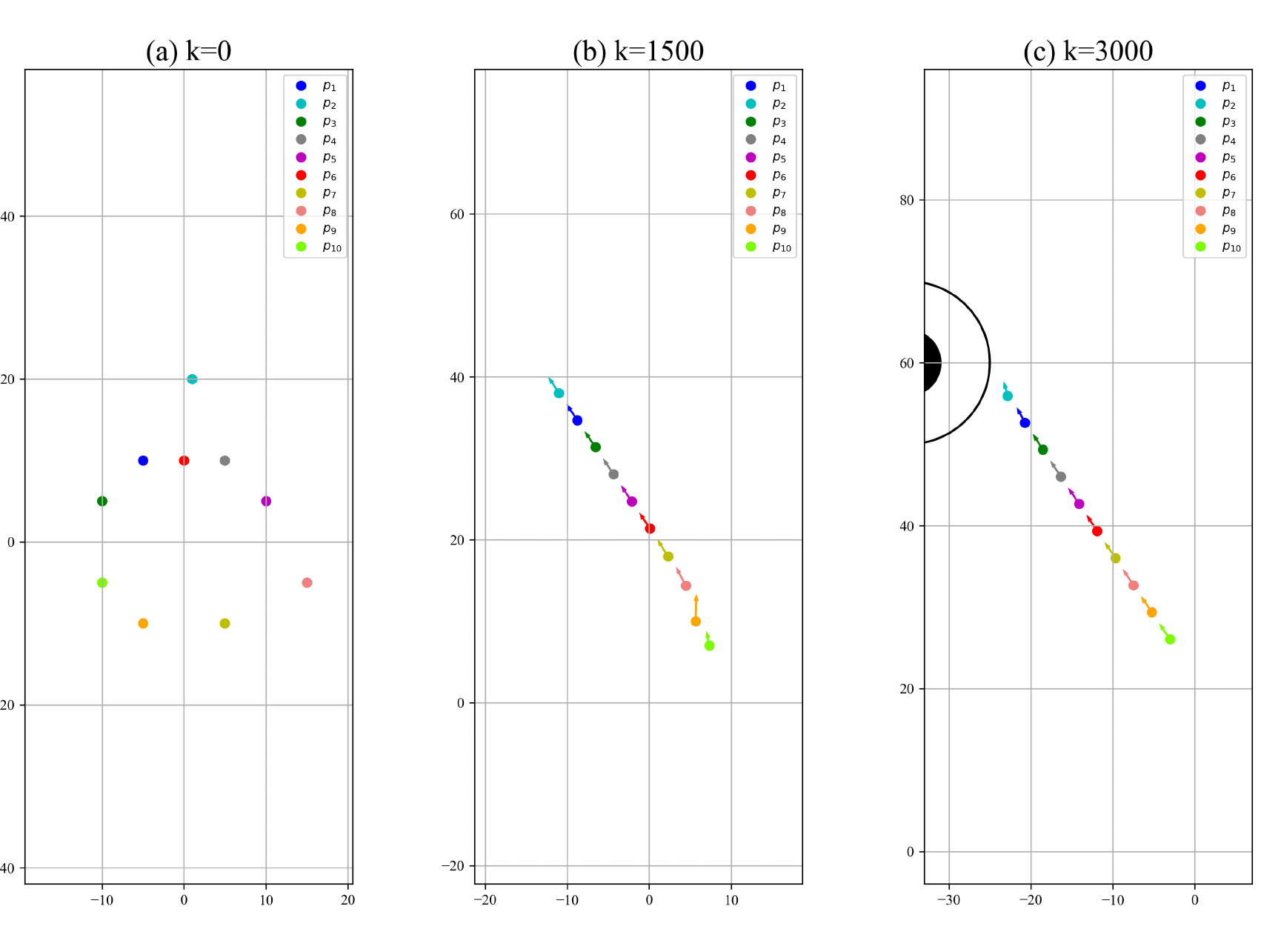}}
    \end{minipage}%
} 
\subfigure{
    \begin{minipage}[t]{0.48\linewidth}
    \centering
    \scalebox{0.4}{\includegraphics[viewport=60 250 550 620, clip]{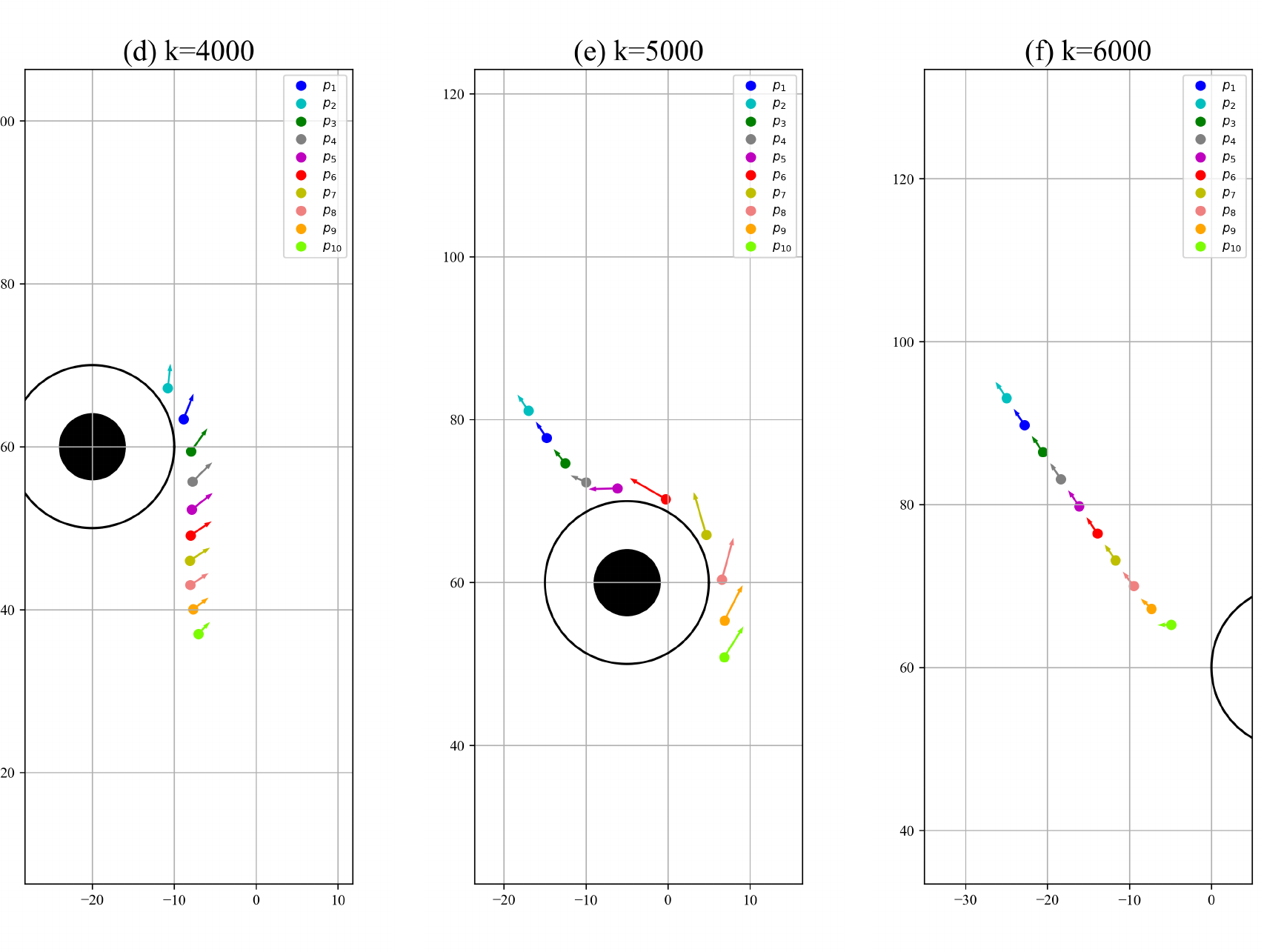}}
    \end{minipage}%
} 
\subfigure{
    \begin{minipage}[t]{0.48\linewidth}
    \centering
    \scalebox{0.4}{\includegraphics[viewport=60 250 550 620, clip]{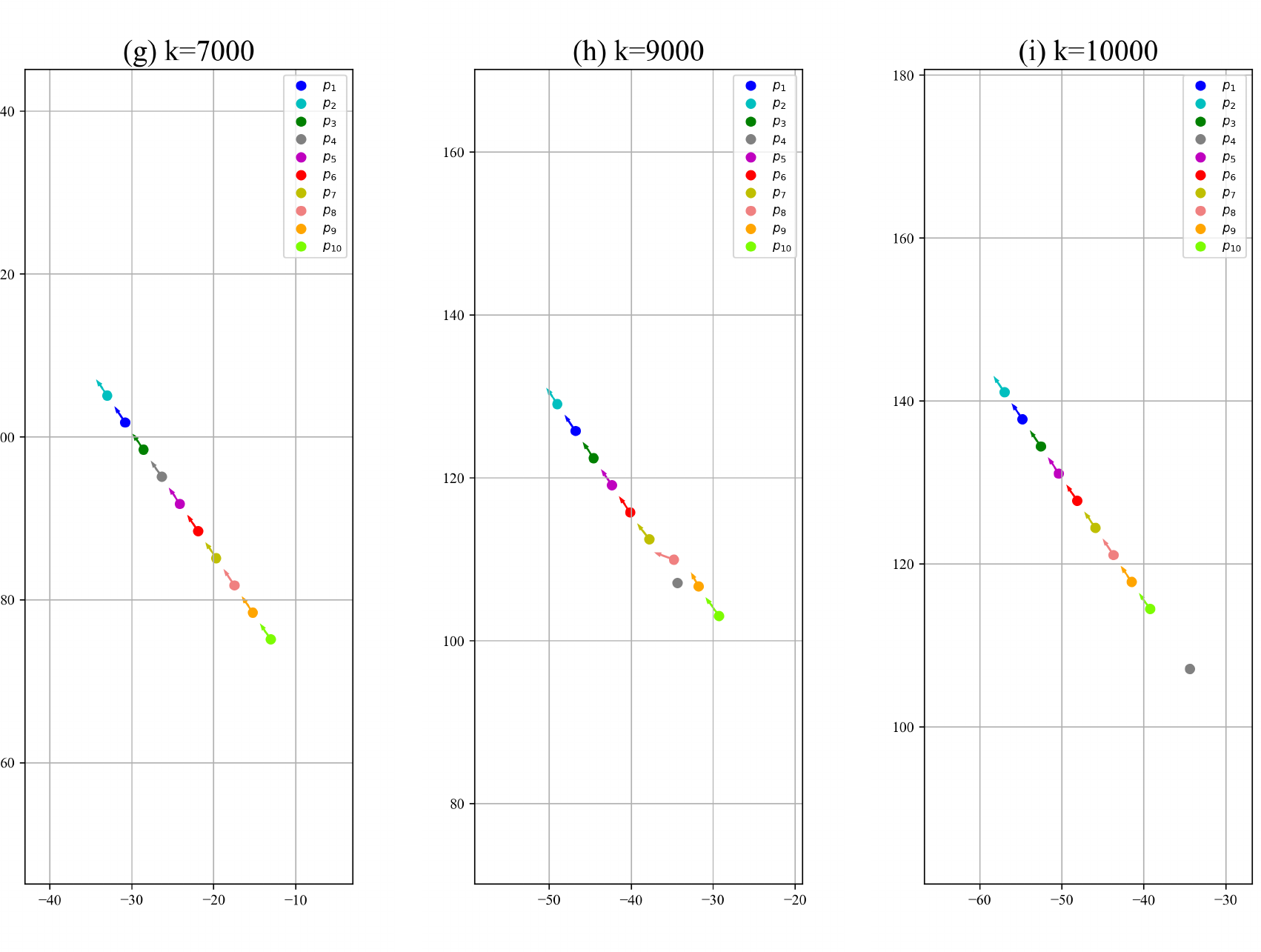}}
    \end{minipage}%
} 
\subfigure{
    \begin{minipage}[t]{0.48\linewidth}
    \centering
    \scalebox{0.4}{\includegraphics[viewport=60 250 550 620, clip]{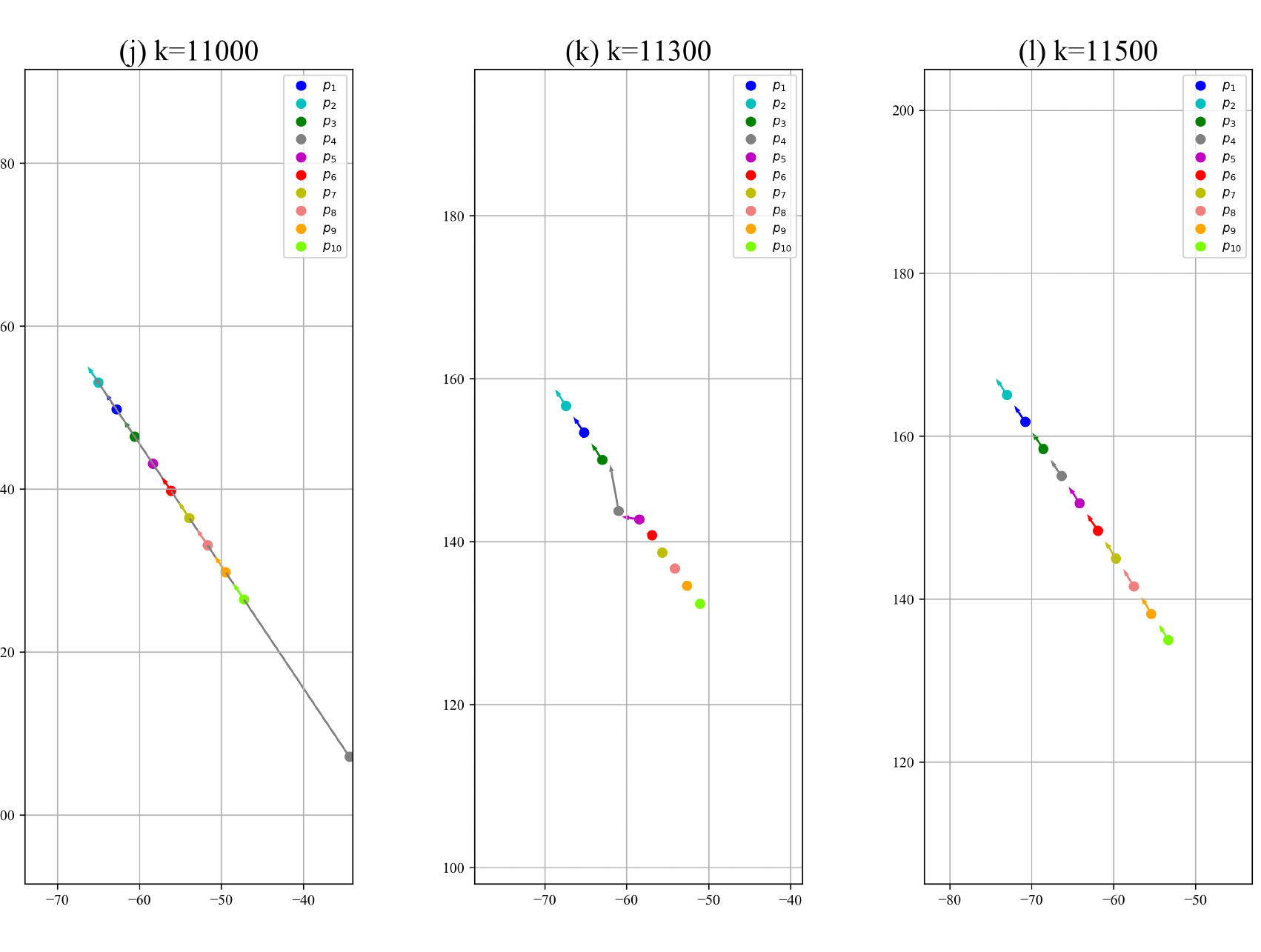}}
    \end{minipage}%
} 

\end{center}

\caption{Robots' and mobile obstacle's positions during the whole simulation process
by Python simulation.}\label{py-4p}

\end{figure*}

Fig. \ref{py-4p} shows the robots' and mobile obstacle's positions during the whole simulation process. It can be seen that the robot
swarm has firstly managed to form a marching line, and then successfully bypassed the mobile obstacle.
When robot 4 fails, other robots consider robot 4 as an immobile obstacle and has re-formed a new marching line.
 After robot 4 gets back to normal, it rejoins the marching line, which happens to be at the same position before the
 event of failure.

\begin{figure*}
\begin{center}

\subfigure{
    \begin{minipage}[t]{0.48\linewidth}
    \centering
    \scalebox{0.3}{\includegraphics[viewport=20 0 680 330, clip]{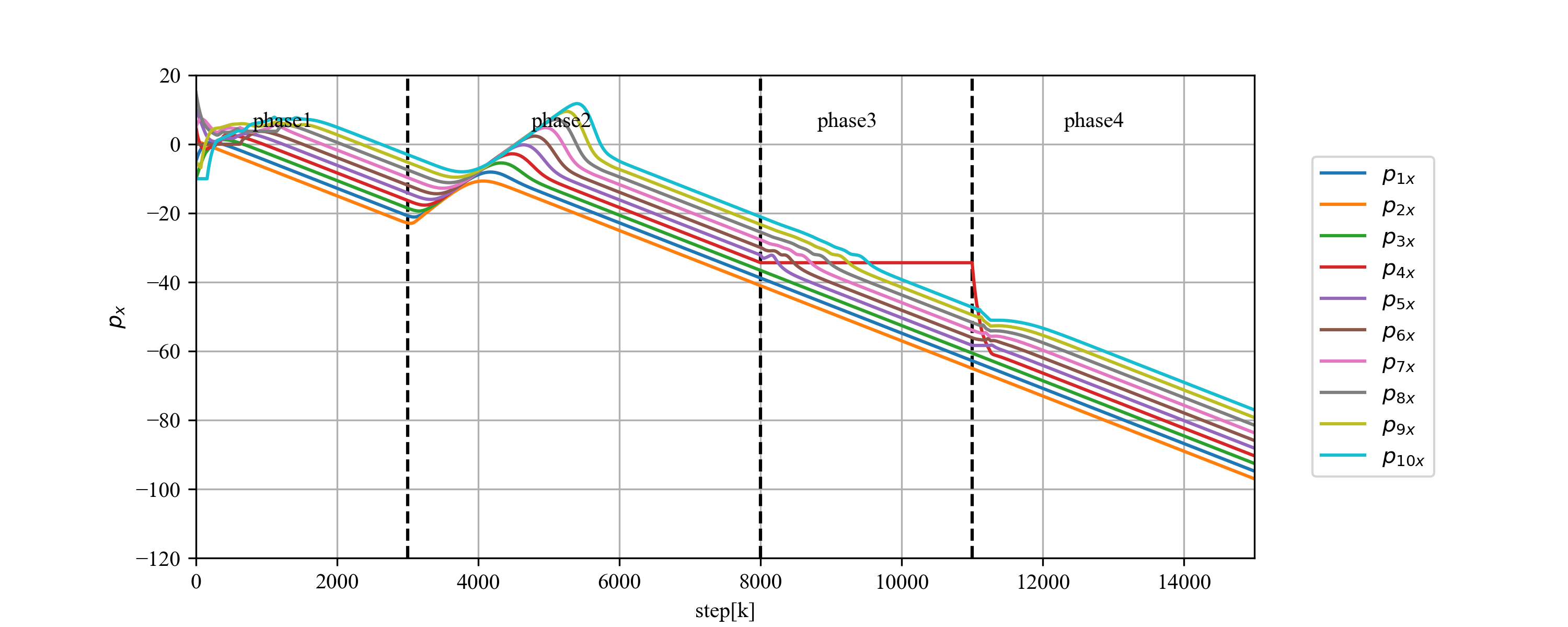}}
    \end{minipage}%
} 
\subfigure{
    \begin{minipage}[t]{0.48\linewidth}
    \centering
    \scalebox{0.3}{\includegraphics[viewport=20 0 680 330, clip]{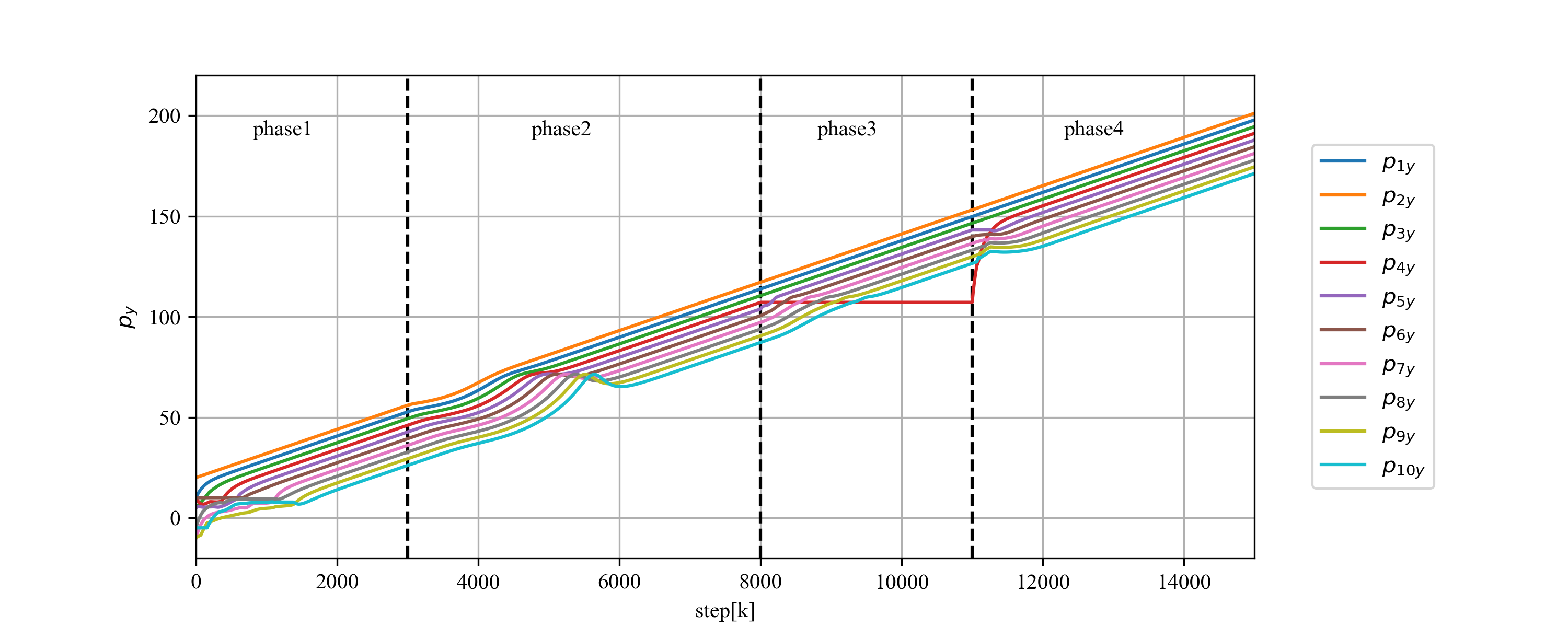}}
    \end{minipage}%
} 

\end{center}

\caption{Robots' positions during the whole process by Python simulation.}\label{py-x}

\end{figure*}

\begin{figure*}
\begin{center}

\subfigure{
    \begin{minipage}[t]{0.48\linewidth}
    \centering
    \scalebox{0.3}{\includegraphics[viewport=20 0 680 330, clip]{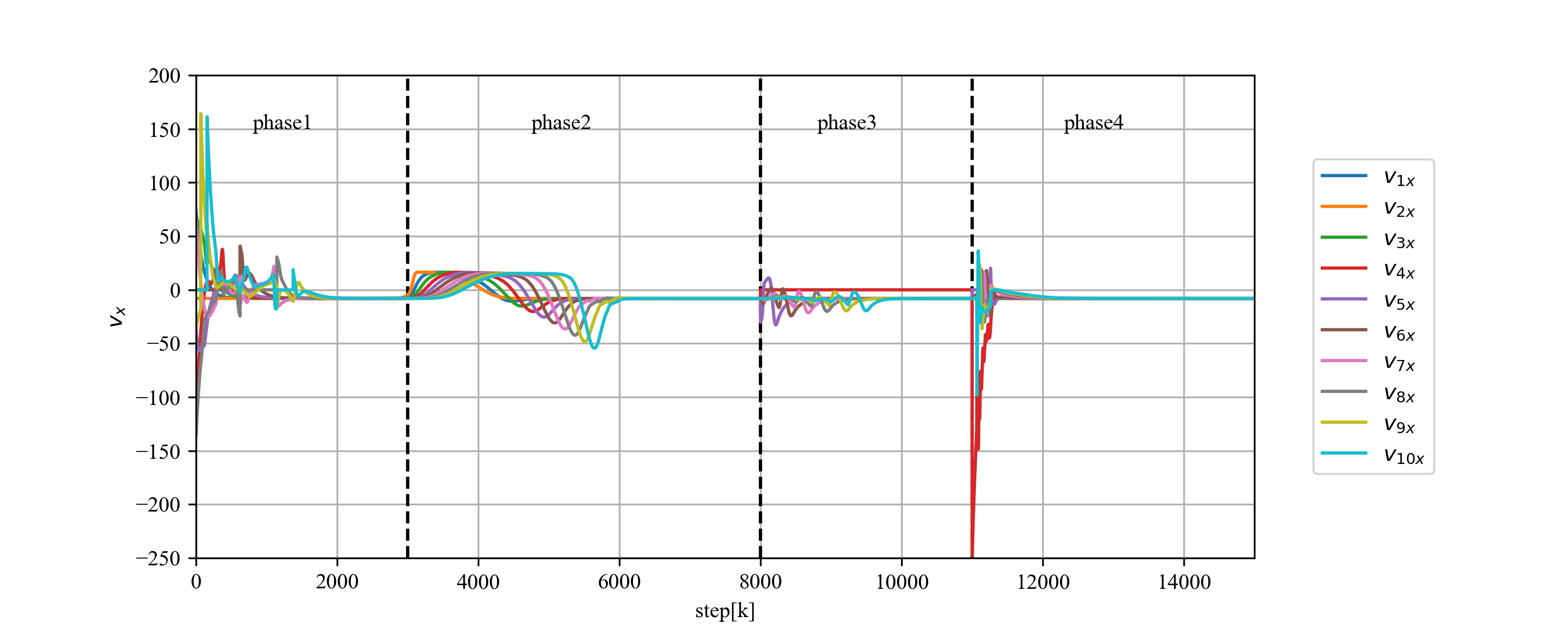}}
    \end{minipage}%
} 
\subfigure{
    \begin{minipage}[t]{0.48\linewidth}
    \centering
    \scalebox{0.3}{\includegraphics[viewport=20 0 680 330, clip]{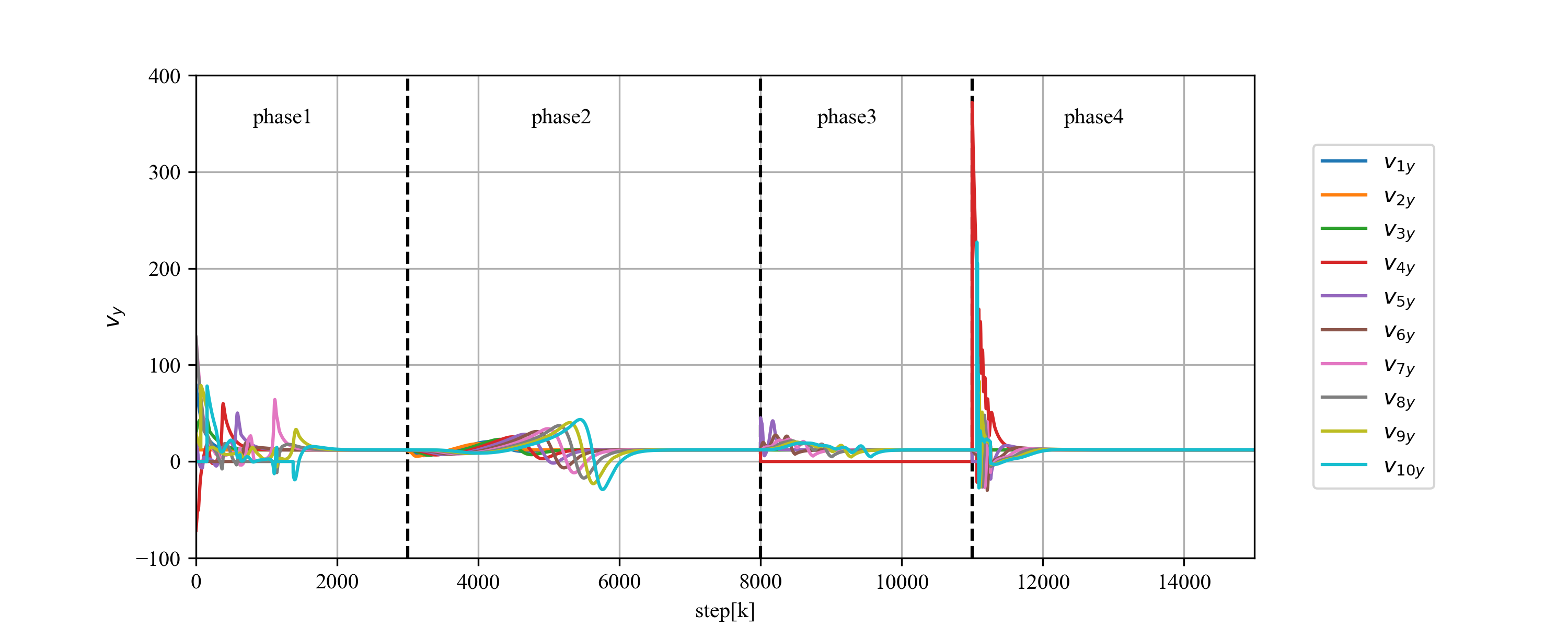}}
    \end{minipage}%
} 

\end{center}

\caption{Robots' velocities during the whole process by Python simulation.}\label{py-v}

\end{figure*}

\begin{figure*}
\begin{center}

\subfigure{
    \begin{minipage}[t]{0.48\linewidth}
    \centering
    \scalebox{0.3}{\includegraphics[viewport=30 0 680 330, clip]{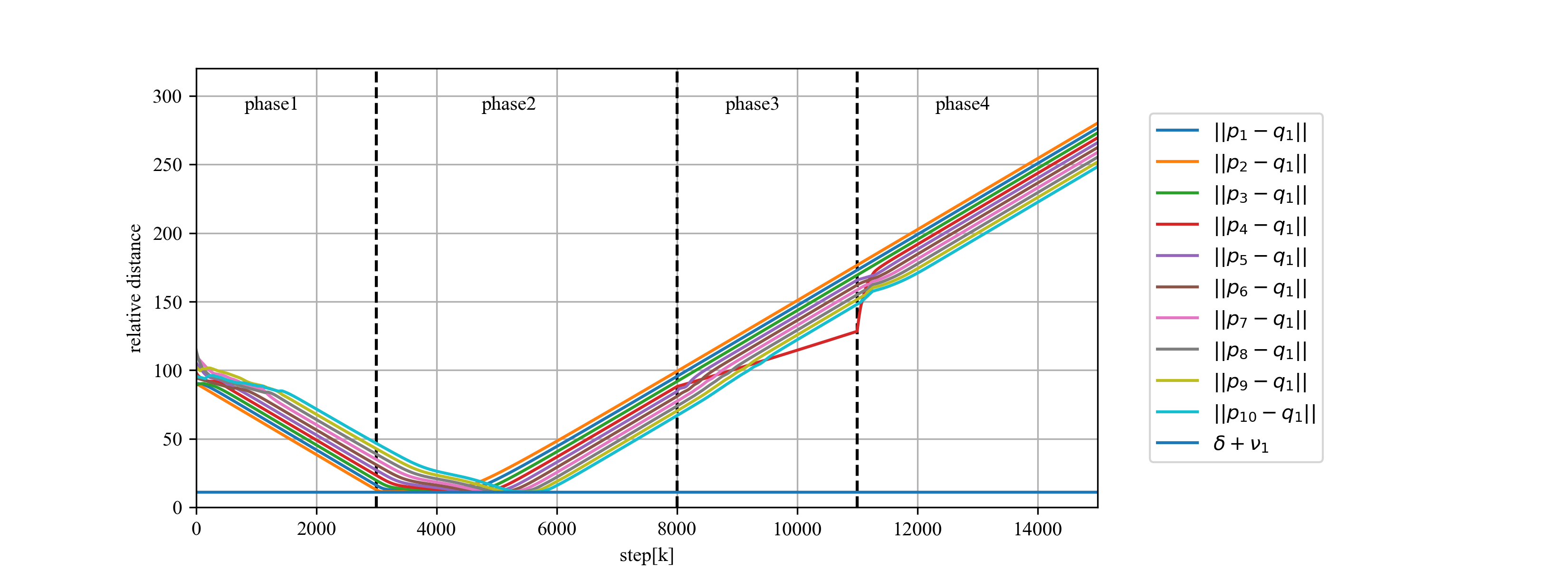}}
    \end{minipage}%
} 
\subfigure{
    \begin{minipage}[t]{0.48\linewidth}
    \centering
    \scalebox{0.3}{\includegraphics[viewport=-10 0 700 300, clip]{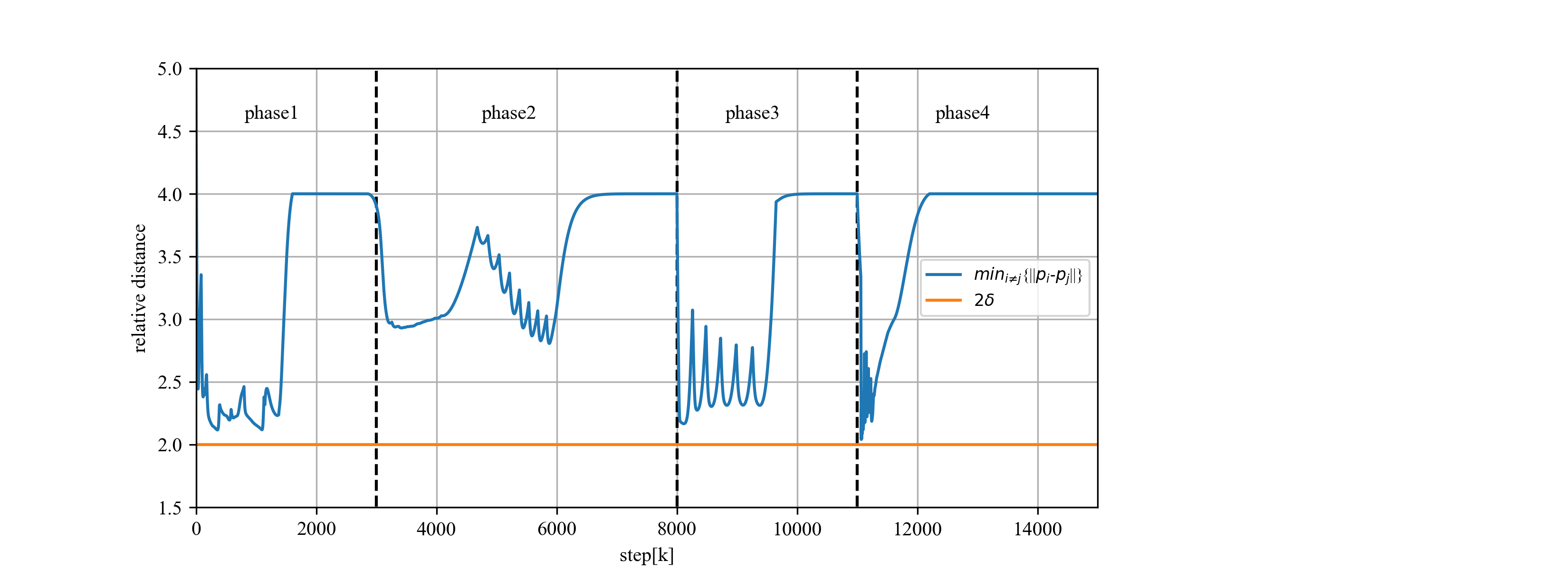}}
    \end{minipage}%
} 

\end{center}

\caption{Relative distance between each robot and the mobile obstacle, and the minimal relative distance between two robots by Python simulation.}\label{py-dis}

\end{figure*}

Figs. \ref{py-x} and \ref{py-v} show the time profiles of the robots' positions and velocities, respectively. It can be seen that the line marching objective has been successfully achieved. Fig. \ref{py-dis} shows the relative distance between each robot and the mobile obstacle, and the minimal relative distance between two robots, which indicates no collision as expected.

\section{Application to the unicycle model}\label{sec.apptoros}

In this section, we consider the more practical situation
 where the robot is described by the unicycle model, and the numerical
simulation is performed on the ROS platform.
In particular, for $i\in\underline{N}$, the kinematic equations of the $i$th robot are given by
\begin{equation}\label{unidyn}
\begin{aligned}
  \dot{p}_i(t)&=\upsilon_i(t)\left(
                       \begin{array}{c}
                         \cos\vartheta_i(t) \\
                         \sin\vartheta_i(t) \\
                       \end{array}
                     \right)\\
    \dot{\vartheta}_i(t)&=\varpi_i(t)
\end{aligned}
\end{equation}
where $p_i(t)\in \mathbb{R}^2$, $\vartheta_i(t)\in \mathbb{R}$ denote the position and heading angle of the $i$th robot, respectively,
and $\upsilon_i(t)\in \mathbb{R}$, $\varpi_i(t)\in \mathbb{R}$ denote the linear and angular velocity, respectively.

Let $l_i$ denote the distance between the wheel pair of the $i$th robot, and then
\begin{equation}\label{wheelspeed}
  \begin{aligned}
    \upsilon_{il}(t)&= \upsilon_i(t)-\frac{1}{2}l_i\varpi_i(t)\\
    \upsilon_{ir}(t)&= \upsilon_i(t)+\frac{1}{2}l_i\varpi_i(t)
  \end{aligned}
\end{equation}
denote the speeds of left wheel and the right wheel of the robot, respectively.
Since the system \eqref{unidyn} is nonlinear, the proposed line marching algorithm cannot be applied directly.
To tackle this issue, we perform the following coordinate transformation as in \cite{renalkins}:
\begin{equation}\label{coordinatetrans}
  \bar{p}_i(t)=p_i(t)+d_i\left(
                       \begin{array}{c}
                         \cos\vartheta_i(t) \\
                         \sin\vartheta_i(t) \\
                       \end{array}
                     \right)
\end{equation}
where $d_i>0$ is some small number. Then it follows that
\begin{equation*}
  \begin{aligned}
    \dot{\bar{p}}_i(t)&=\dot{p}_i(t)+d_i\varpi_i(t)\left(
                       \begin{array}{c}
                         -\sin\vartheta_i(t)  \\
                         \cos\vartheta_i(t)\\
                       \end{array}
                     \right)\\
    &=\upsilon_i(t)\left(
                       \begin{array}{c}
                         \cos\vartheta_i(t) \\
                         \sin\vartheta_i(t) \\
                       \end{array}
                     \right)+d_i\varpi_i(t)\left(
                       \begin{array}{c}
                         -\sin\vartheta_i(t)  \\
                         \cos\vartheta_i(t)\\
                       \end{array}
                     \right)\\
  &=\left(
      \begin{array}{cc}
        \cos\vartheta_i(t) & -d_i\sin\vartheta_i(t) \\
        \sin\vartheta_i(t) & d_i\cos\vartheta_i(t) \\
      \end{array}
    \right)\left(
             \begin{array}{c}
               \upsilon_i(t) \\
               \varpi_i(t) \\
             \end{array}
           \right).
  \end{aligned}
\end{equation*}
Let
\begin{equation*}
  \Theta_i(t)=\left(
      \begin{array}{cc}
        \cos\vartheta_i(t) & -d_i\sin\vartheta_i(t) \\
        \sin\vartheta_i(t) & d_i\cos\vartheta_i(t) \\
      \end{array}
    \right)
\end{equation*}
and
\begin{equation*}
  \bar{v}_i(t)=\Theta_i(t)\left(
             \begin{array}{c}
               \upsilon_i(t) \\
               \varpi_i(t) \\
             \end{array}
           \right).
\end{equation*}
Then, we have
\begin{equation}\label{transed}
  \dot{\bar{p}}_i(t)=\bar{v}_i(t).
\end{equation}
Moreover, it can be verified that for all $t\geq 0$, $\Theta_i(t)$ is nonsingular whose inverse is given by
\begin{equation*}
  \Theta_i(t)^{-1}=\left(
       \begin{array}{cc}
         \cos\vartheta_i(t) & \sin\vartheta_i(t) \\
         -\sin\vartheta_i(t)/d_i & \cos\vartheta_i(t)/d_i \\
       \end{array}
     \right).
\end{equation*}

By using the coordinate transformation \eqref{coordinatetrans}, the nonlinear system \eqref{unidyn} is transformed
into a linear kinematic system \eqref{transed}, which is in the same form of \eqref{robotdyn} and thus the
proposed algorithm is applicable. After obtaining $\bar{v}_i(t)$,
the linear and angular velocity inputs to system \eqref{unidyn} are given by
\begin{equation}\label{}
  \left(
             \begin{array}{c}
               \upsilon_i(t) \\
               \varpi_i(t) \\
             \end{array}
           \right)=\Theta_i(t)^{-1}\bar{v}_i(t)
\end{equation}
and then the speeds of left and right wheel of the robot can be determined by
equation \eqref{wheelspeed}.

Note that in ROS platform,
the position measurements of the robots are subject to the Gaussian noise.
As a result, the co-leader situation would almost be impossible.
Thus, Algorithm 1
can be further simplified into Algorithm 3.
For ROS simulation, the robot moves continuously with respect to time.
While, to implement the algorithm in a decentralized way, it takes time
for inter-robot communication. Thus, the desired velocities for the robot swarm
will be updated periodically with $T$ denoting the period.

\begin{algorithm}[h]\label{ag3}
\caption{Line Marching Algorithm (ROS)} 
\hspace*{0.02in} {\bf Input:} 
$p_1(t),\dots,p_N(t)$.\\
\hspace*{0.02in} {\bf Output:} 
$v_1(t),\dots,v_N(t)$.
\begin{algorithmic}[1]
\For{$i=1$, $i\leq N$, $i=i+1$}
\State set $\Delta_i=0$, $\Phi_i=0$, $\Lambda_i=1$.
\State set $v_i(t)=0$.
\EndFor
\For{$i=1$, $i\leq N$, $i=i+1$}
\If{$\Gamma_i=1$}
\For{$j=1$, $j\leq N$, $j\neq i$, $j=j+1$ \& $\Phi_i=0$ }
\If {$\langle p_j(t)-p_i(t),e_l\rangle>0$ \& $\Gamma_j=1$}
\State set $\Lambda_i=0$.
\If {$\Delta_j=0$}
\State $v_i(t)=v_{i,lm}^j(t)+v_{i,ca}(t)$,
\State set $\Delta_j=1$, $\Phi_i=1$.
\EndIf
\EndIf
\EndFor
\If {$\Lambda_i=1$}
\State $v_i(t)=v_l+v_{i,ca}(t)$.
\EndIf
\EndIf
\EndFor
\State \Return result
\end{algorithmic}
\end{algorithm}

Different from the cases studied in Sections \ref{sec.apptocon} and \ref{sec.apptodis},
in ROS simulation, the linear and angular velocities of the robots are
subject to saturation.
For $i\in \underline{N}$, suppose
 $|\upsilon_i(t)|\leq \upsilon_{\max}$, $|\varpi_i(t)|\leq \omega_{\max}$.

Similar to the cases studied in Sections \ref{sec.apptocon} and \ref{sec.apptodis}, we consider
a swarm of 10 robots. Let $T=0.02\mathrm{s}$.
The parameters for the line marching are given by $v_l=(-0.25,0.433)^T$, $\rho=2$.
For simplicity, the minimal inter-robot safety distance is set to be $\delta_i\triangleq\delta=0.35$.
Suppose $\upsilon_{\max} = 1\mathrm{m/s}$, $\omega_{\max} = 2\mathrm{rad/s}$.
The control parameters are selected to be $\kappa_1=2.5$, $\kappa_2=10$,
$a_i=10$, $\omega_i=0.3$, $\alpha=1$, $\beta=0.5$, $d_i = 0.2$.
Suppose there is a mobile obstacle with $\nu_1=2$ and $u_1(t)=(0.2133,0.211)^T$.

The initial positions of the robots are given by
$p_1(0)=(-4,8)$, $p_2(0)=(0,12)$, $p_3(0)=(-8,4)$, $p_4(0)=(4,8)$, $p_5(0)=(8,4)$, $p_6(0)=(0,8)$, $p_7(0)=(4,-8)$, $p_8(0)=(12,-4)$, $p_9(0)=(-4,-8)$, $p_{10}(0)=(-8, -4)$. The initial heading angles are given by $\vartheta_i(0)=0$. The initial position for the
obstacle is given by $q_1(0)=(-30,23)^T$.

Similar to the case studied in Section \ref{sec.apptodis},
the whole simulation process is divided into four phases:
\begin{enumerate}
  \item from $t=0$s to $t=54$s, the robots will form a line from the initial position;
  \item from $t=54$s to $t=130$s, the robots will bypass the mobile obstacle and re-form a marching line;
  \item from $t=130$s to $t=160$s, robot 6 fails;
  \item from $t=160$s to $t=218.8$s, robot 6 gets back to normal and rejoins the robot swarm.
\end{enumerate}

\begin{figure}
\begin{center}
\scalebox{0.3}{\includegraphics[bb=40 150 570 690]{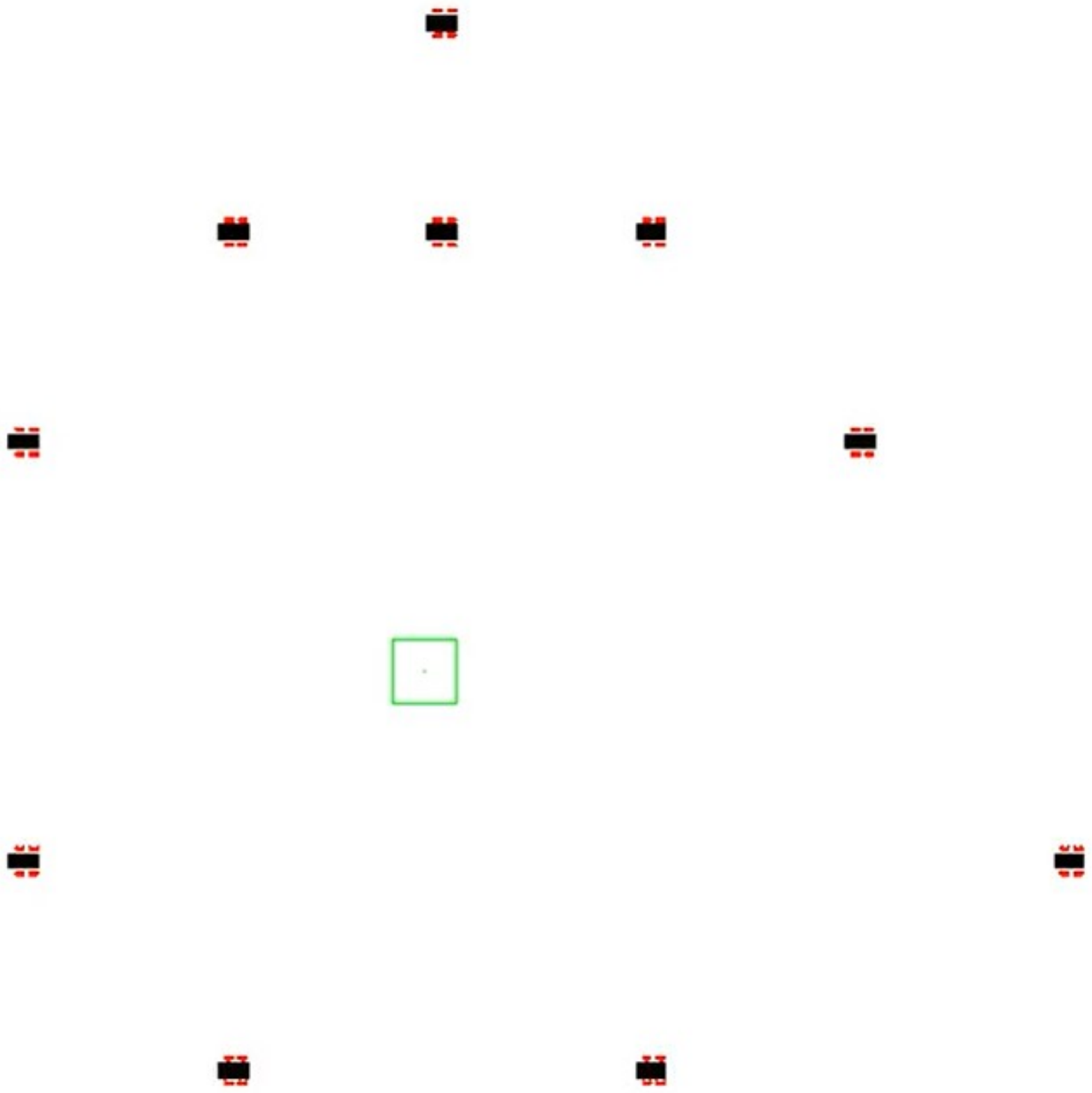}}
\scalebox{0.3}{\includegraphics[bb=40 150 570 690]{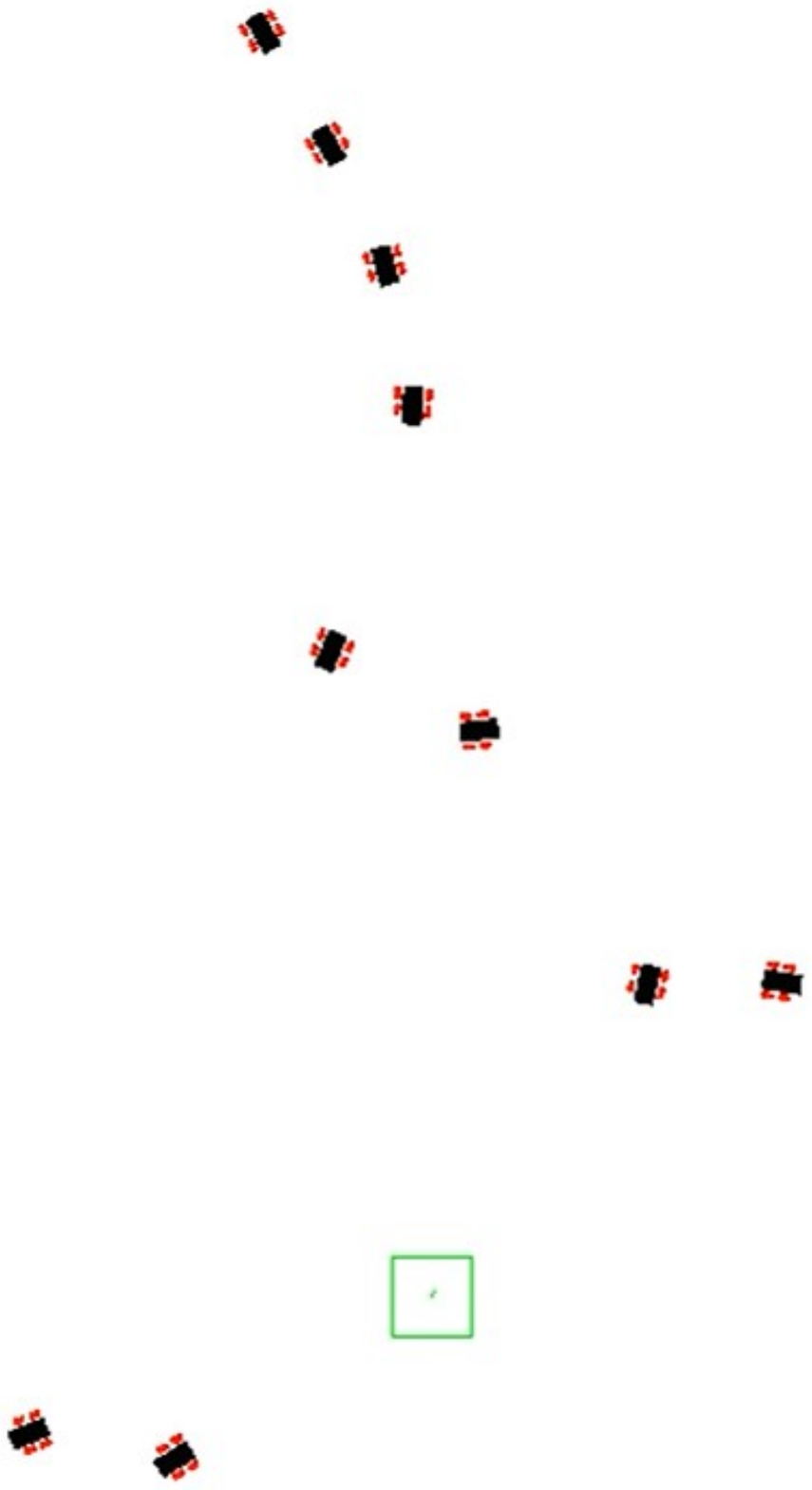}}
\scalebox{0.3}{\includegraphics[bb=40 150 650 850]{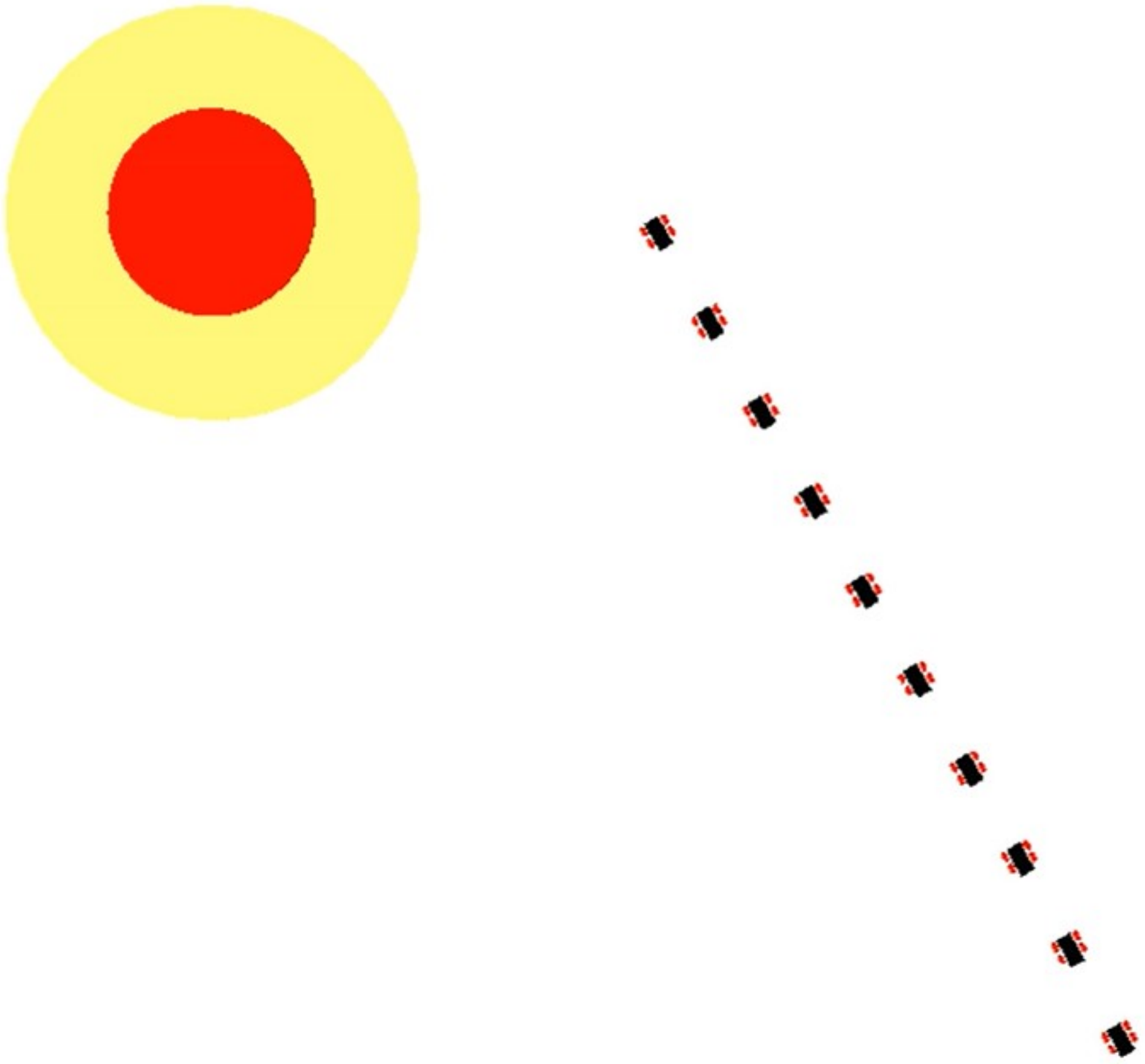}}
\end{center}
\caption{Robots' and mobile obstacle's positions in phase 1 by ROS simulation.}\label{ROS_phase1}
\end{figure}

\begin{figure}
\begin{center}
\scalebox{0.3}{\includegraphics[bb=40 150 570 690]{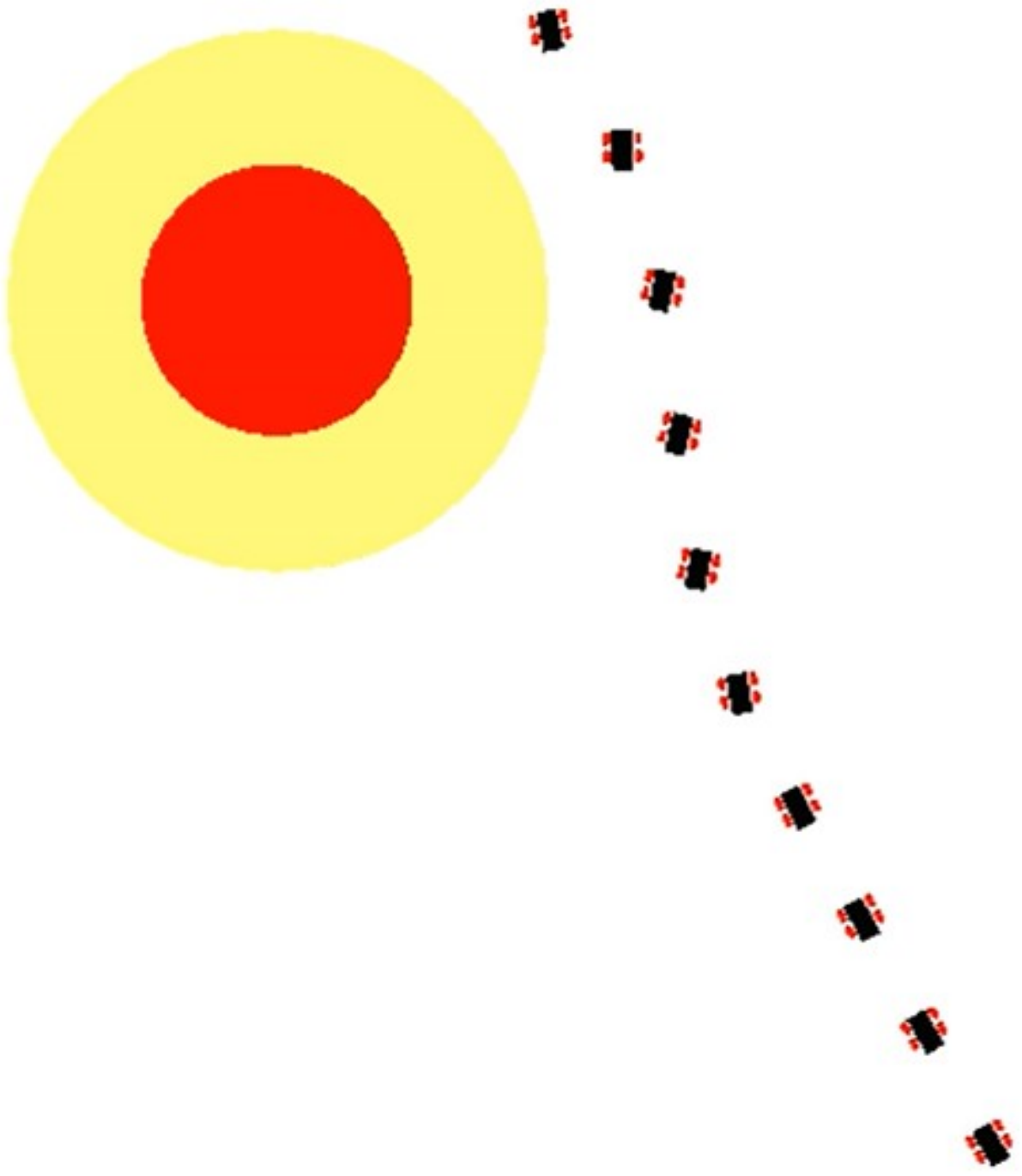}}
\scalebox{0.3}{\includegraphics[bb=40 150 570 690]{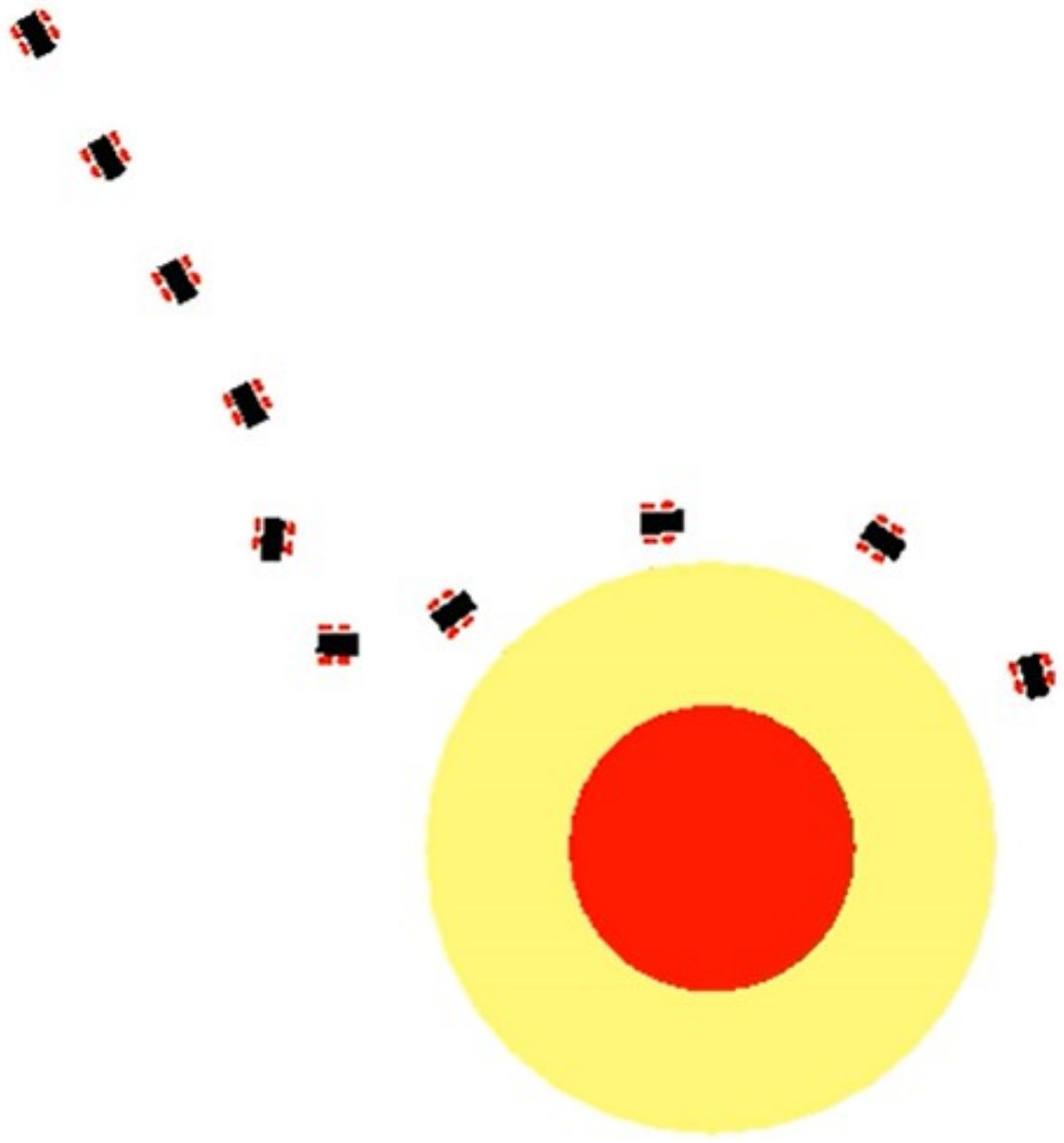}}
\scalebox{0.3}{\includegraphics[bb=40 150 600 800]{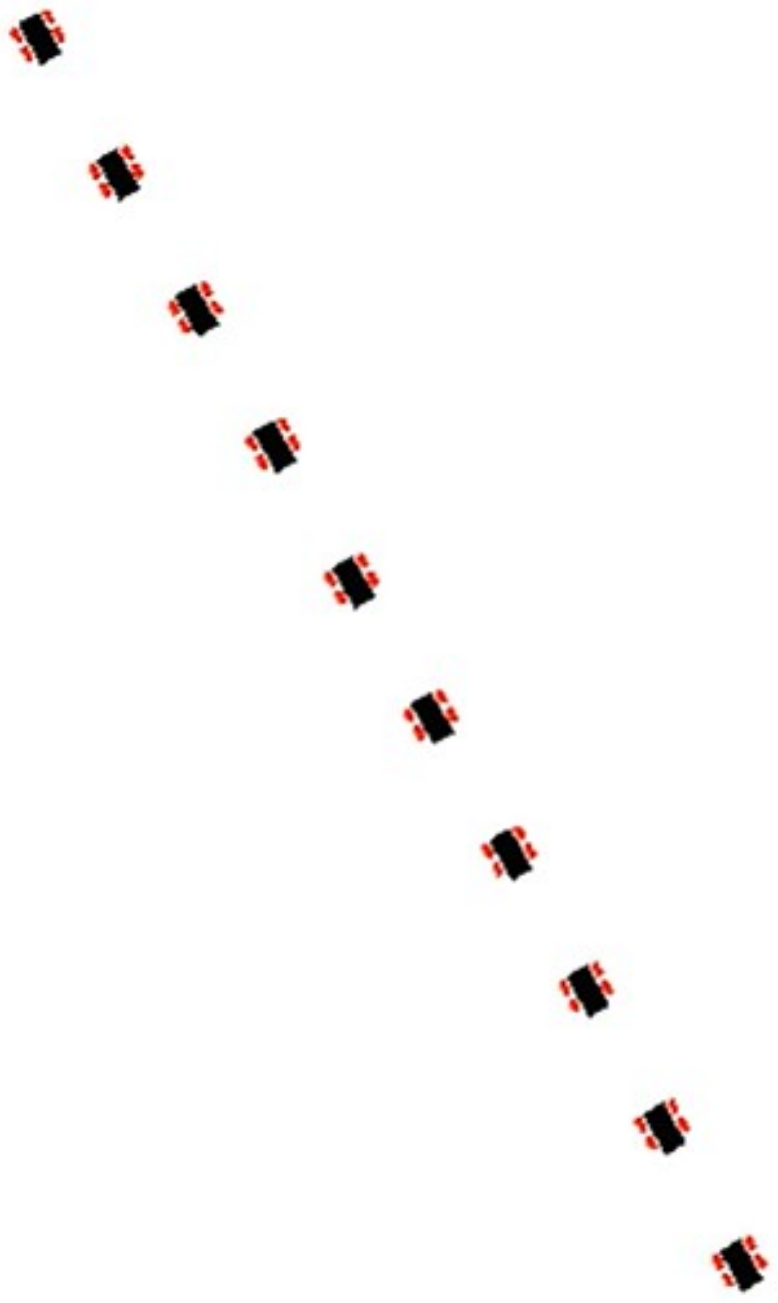}}
\end{center}
\caption{Robots' and mobile obstacle's positions in phase 2 by ROS simulation.}\label{ROS_phase2}
\end{figure}

\begin{figure}
\begin{center}
\scalebox{0.3}{\includegraphics[bb=40 150 570 690]{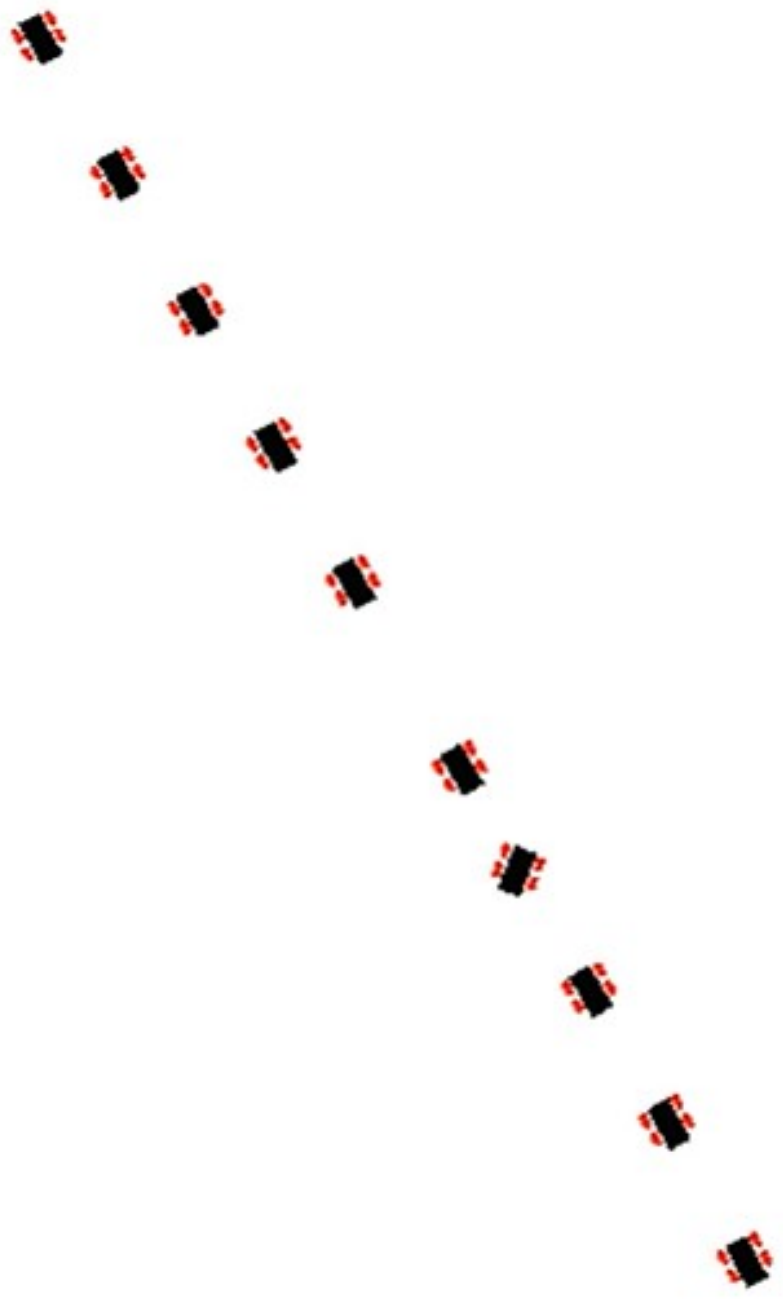}}
\scalebox{0.3}{\includegraphics[bb=40 150 570 690]{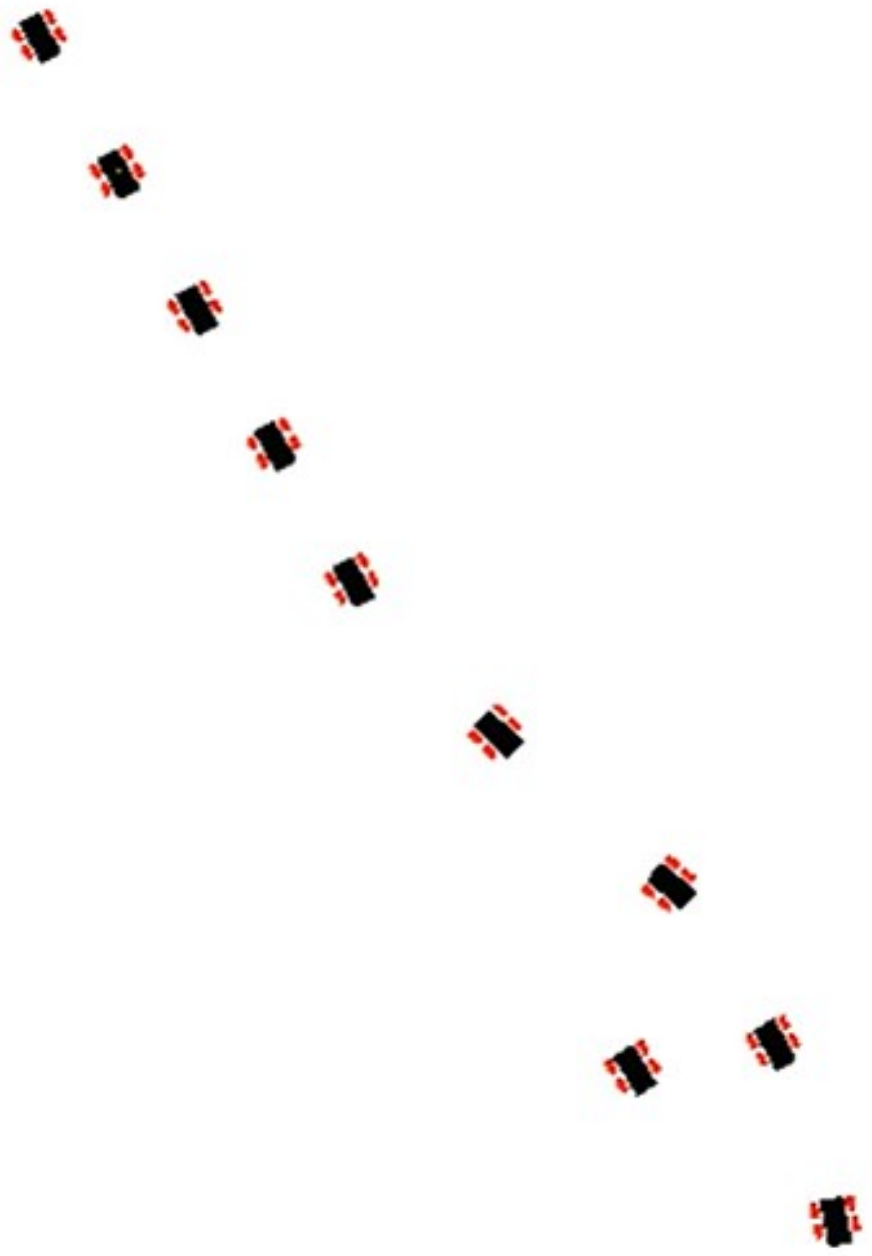}}
\scalebox{0.3}{\includegraphics[bb=40 150 600 800]{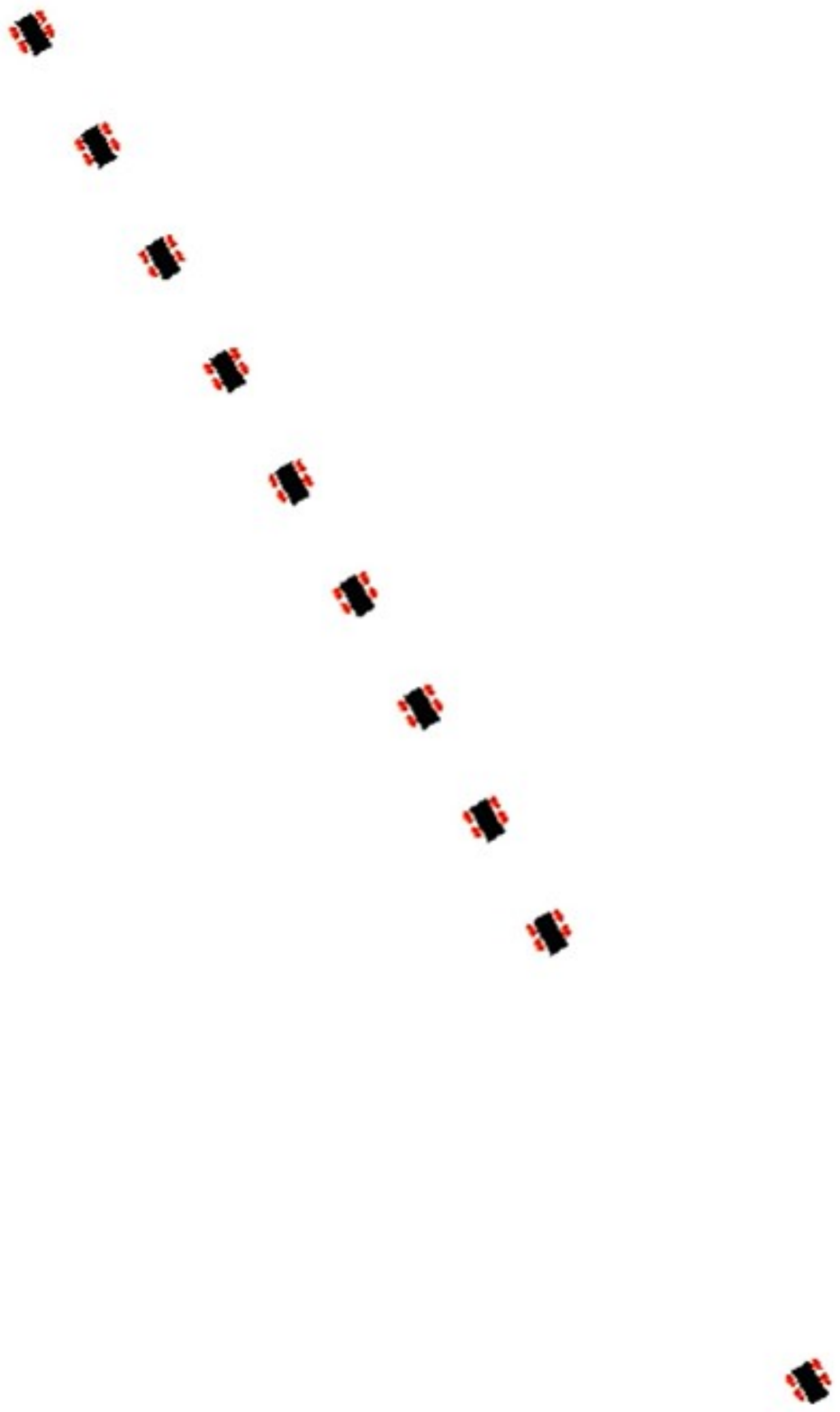}}
\end{center}
\caption{Robots' and mobile obstacle's positions in phase 3 by ROS simulation.}\label{ROS_phase3}
\end{figure}

\begin{figure}
\begin{center}
\scalebox{0.3}{\includegraphics[bb=40 150 570 690]{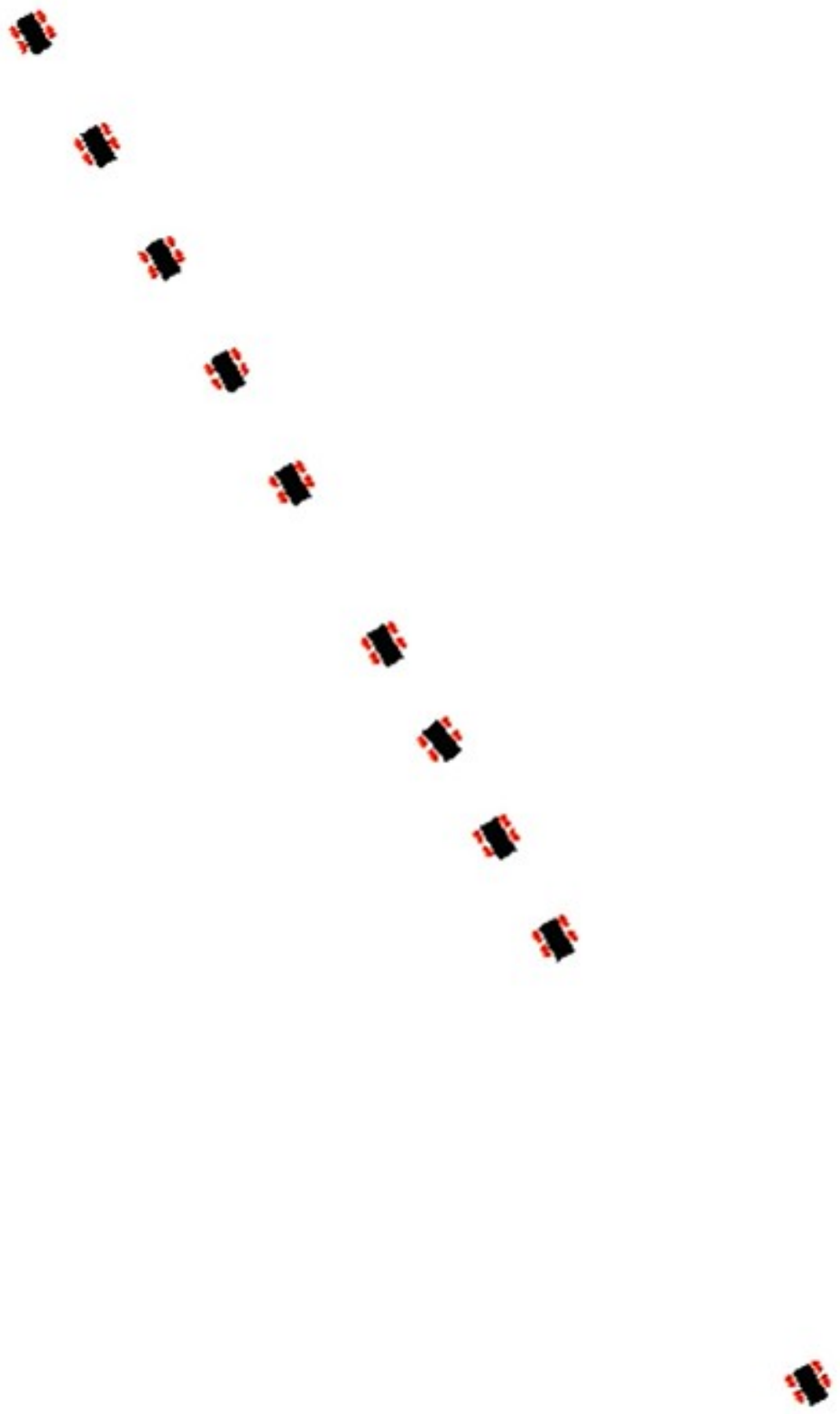}}
\scalebox{0.3}{\includegraphics[bb=40 150 570 690]{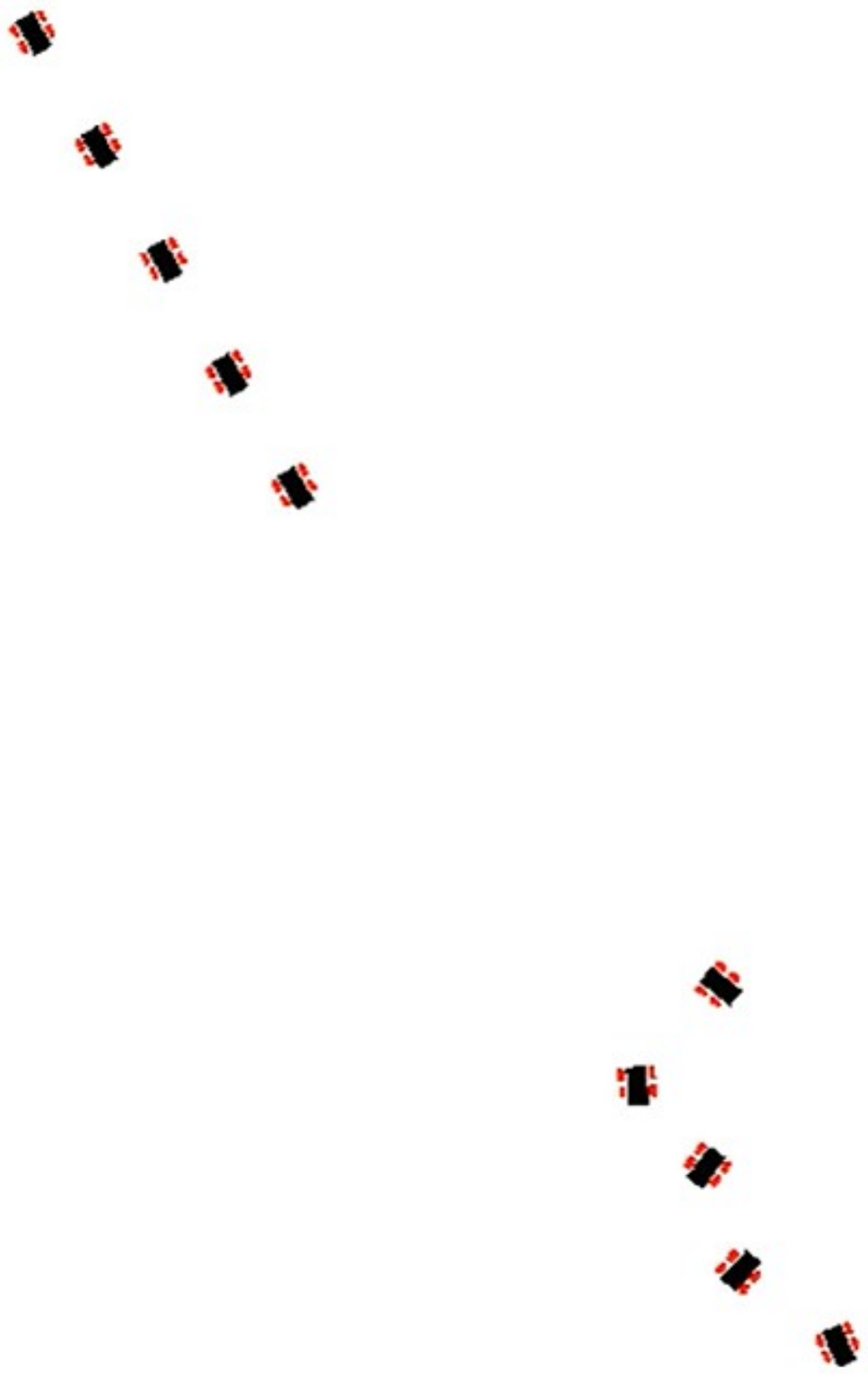}}
\scalebox{0.3}{\includegraphics[bb=40 150 600 800]{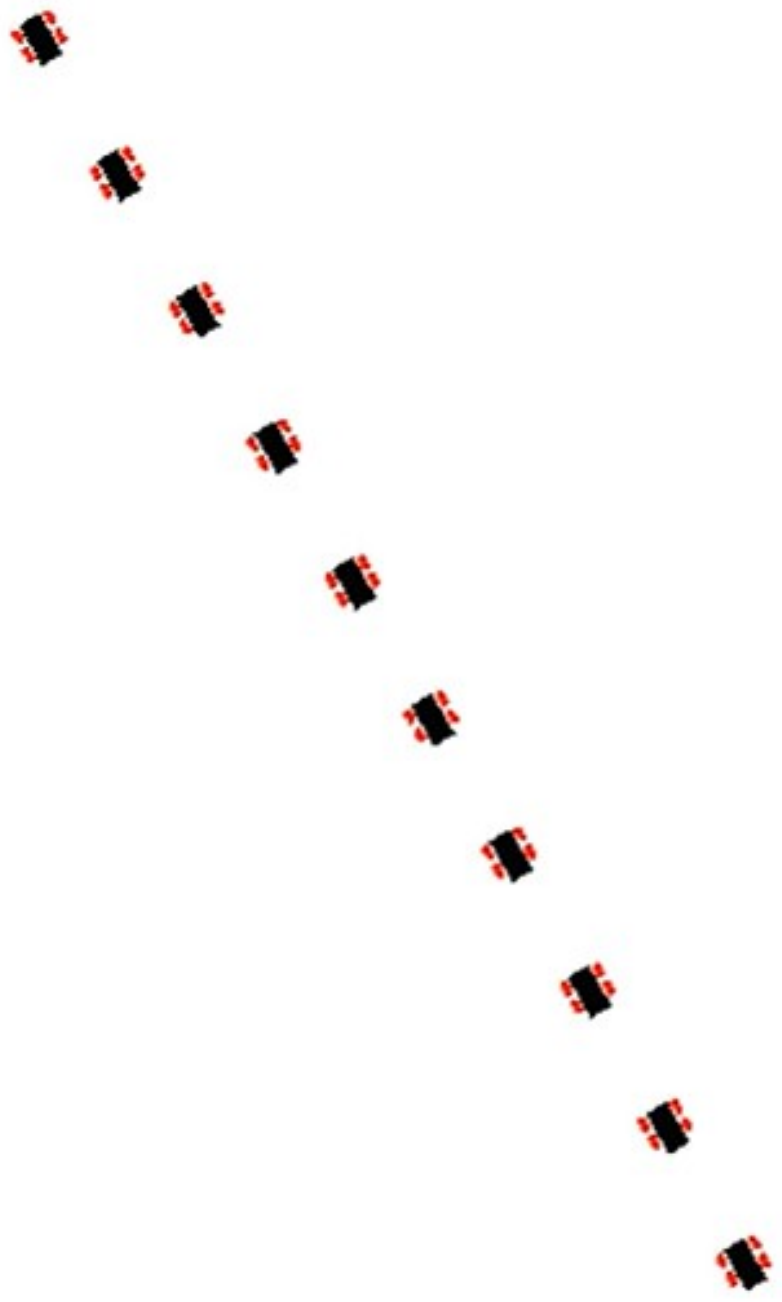}}
\end{center}
\caption{Robots' and mobile obstacle's positions in phase 4 by ROS simulation.}\label{ROS_phase4}
\end{figure}

\begin{figure} 
\begin{center}

\subfigure{
    \begin{minipage}[t]{0.48\linewidth}
    \centering
    \scalebox{0.4}{\includegraphics{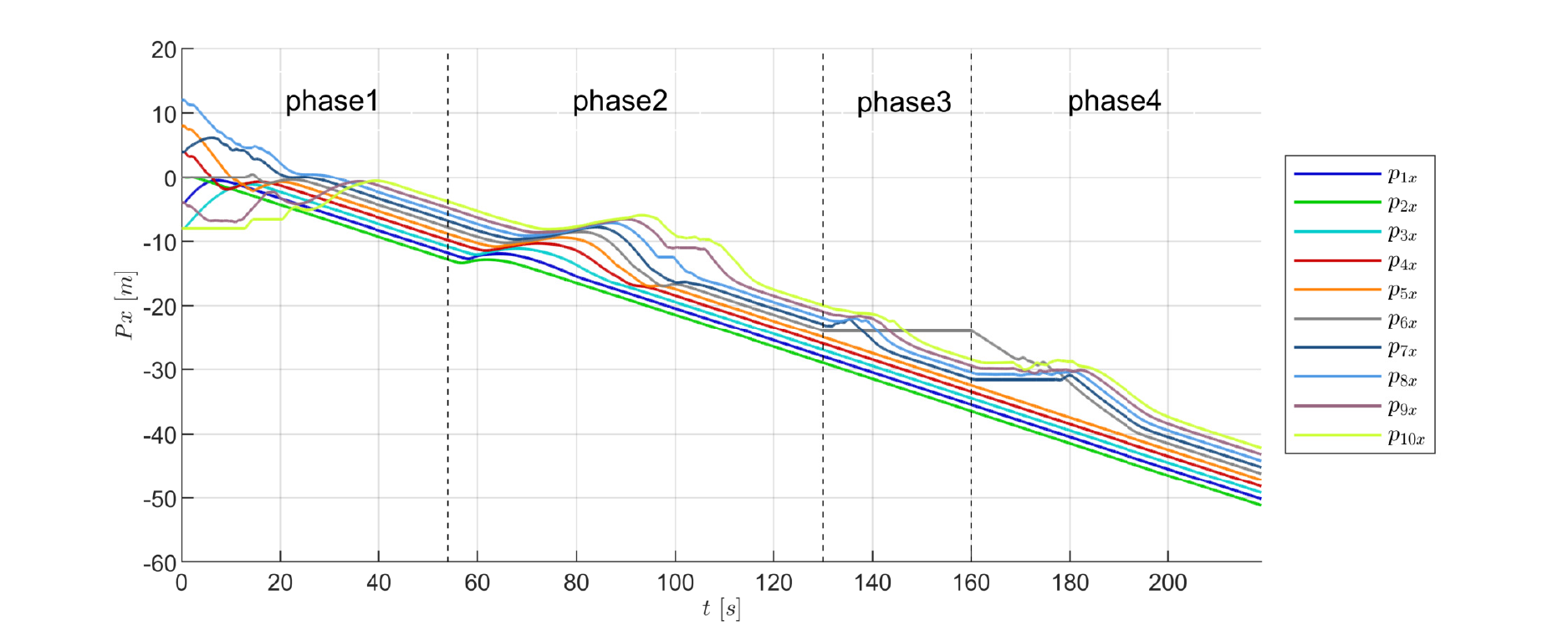}}
    \end{minipage}%
} 
\subfigure{
    \begin{minipage}[t]{0.48\linewidth}
    \centering
    \scalebox{0.4}{\includegraphics{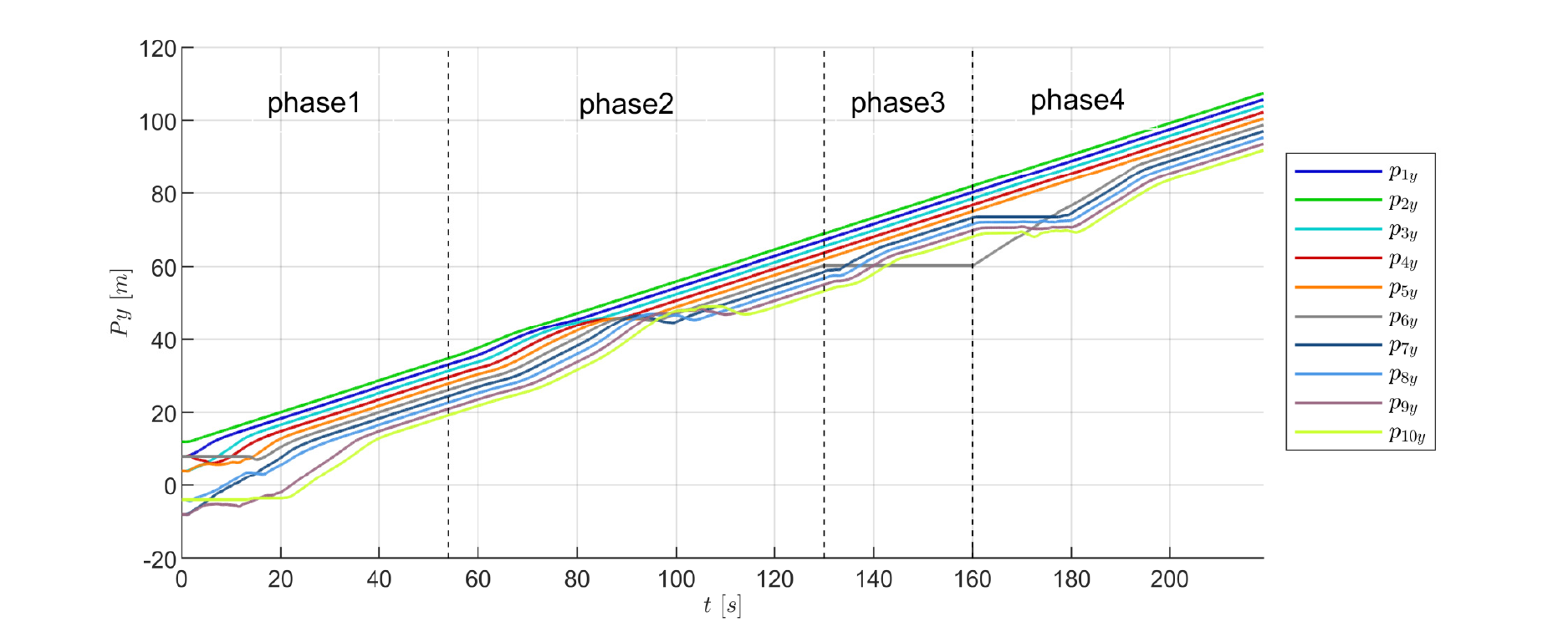}}
    \end{minipage}%
} 

\end{center}
\caption{Robots' positions during the whole process by ROS simulation.}\label{ROS_xyp}
\end{figure}

\begin{figure*} 
\begin{center}

\subfigure{
    \begin{minipage}[t]{0.48\linewidth}
    \centering
    \scalebox{0.4}{\includegraphics{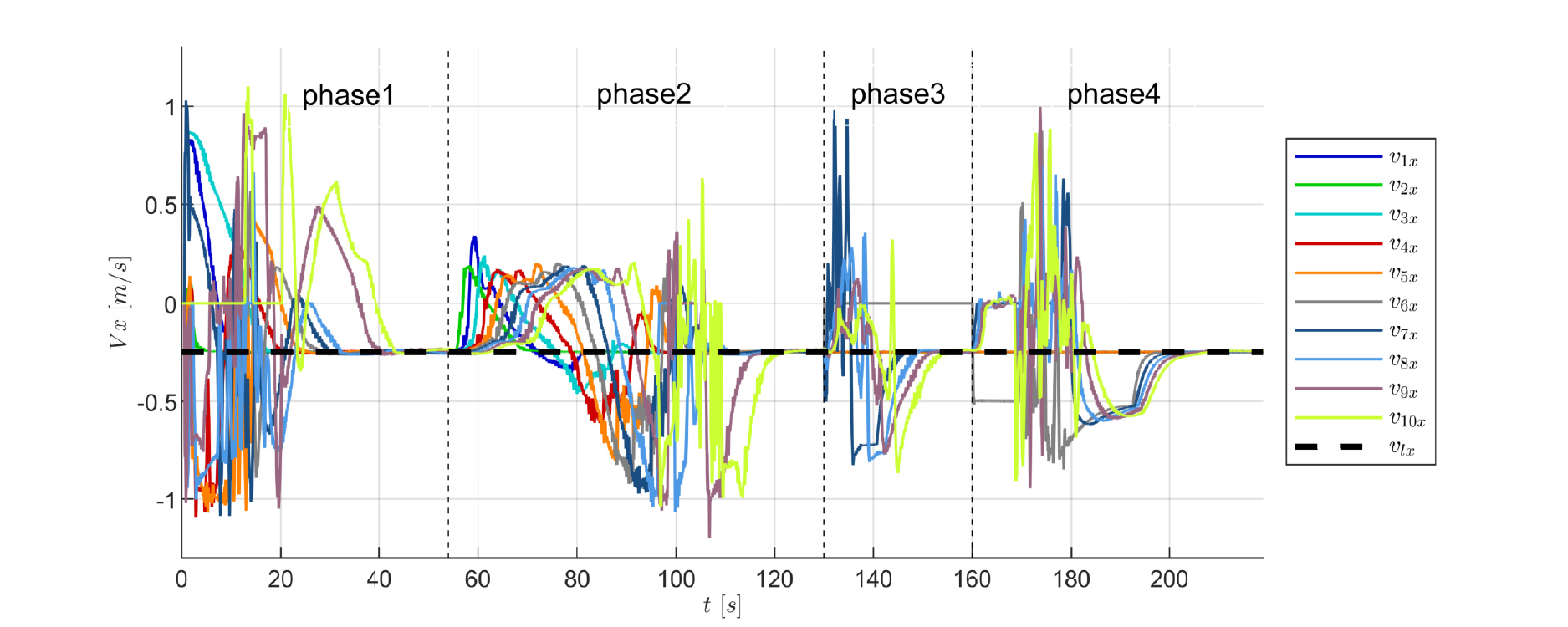}}
    \end{minipage}%
} 
\subfigure{
    \begin{minipage}[t]{0.48\linewidth}
    \centering
    \scalebox{0.4}{\includegraphics{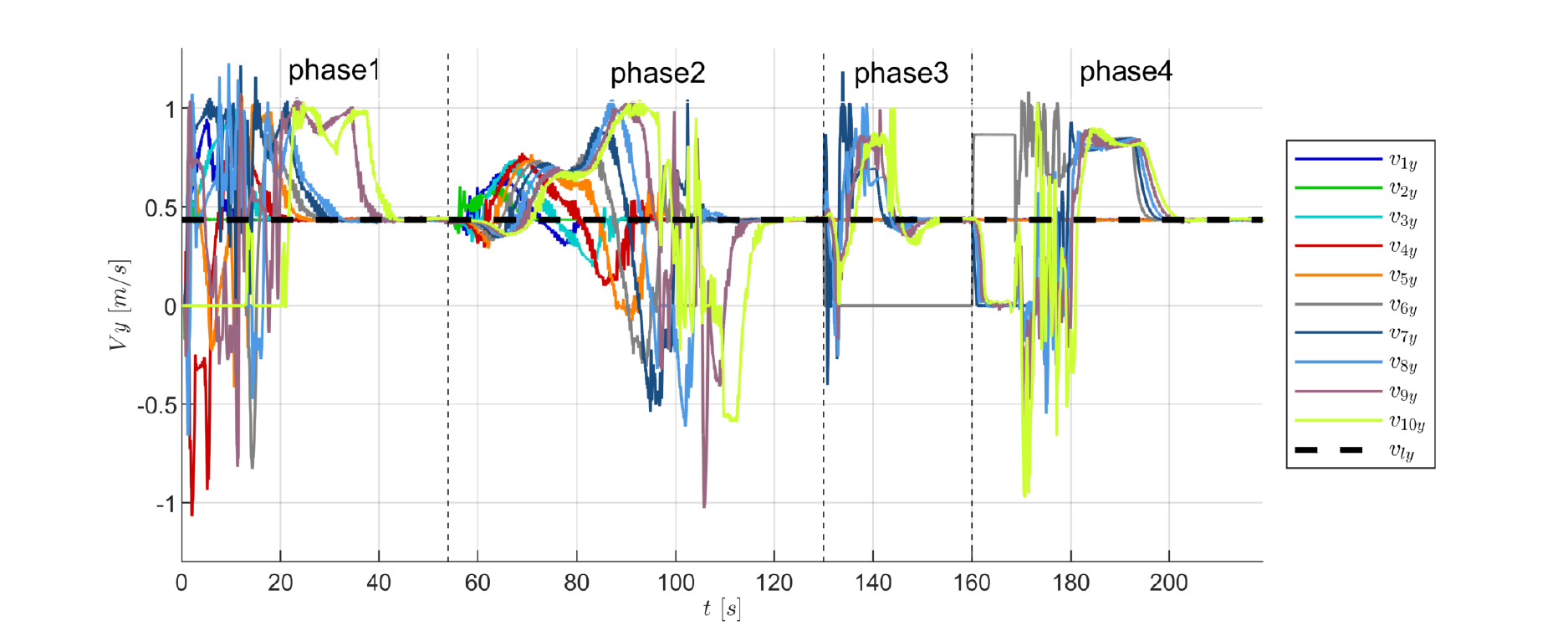}}
    \end{minipage}%
} 

\end{center}
\caption{Robots' velocities during the whole process by ROS simulation.}\label{ROS_xyv}
\end{figure*}

\begin{figure*} 
\begin{center}

\subfigure{
    \begin{minipage}[t]{0.48\linewidth}
    \centering
    \scalebox{0.4}{\includegraphics{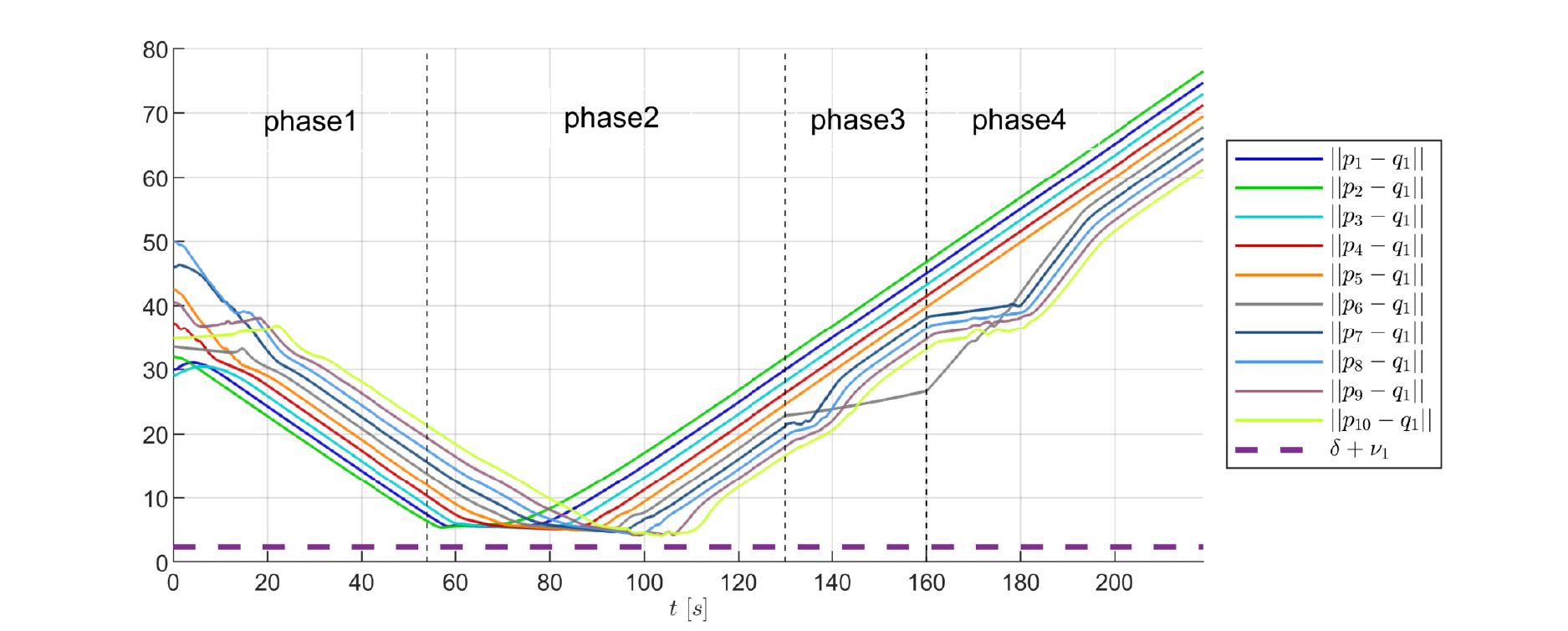}}
    \end{minipage}%
} 
\subfigure{
    \begin{minipage}[t]{0.48\linewidth}
    \centering
    \scalebox{0.4}{\includegraphics{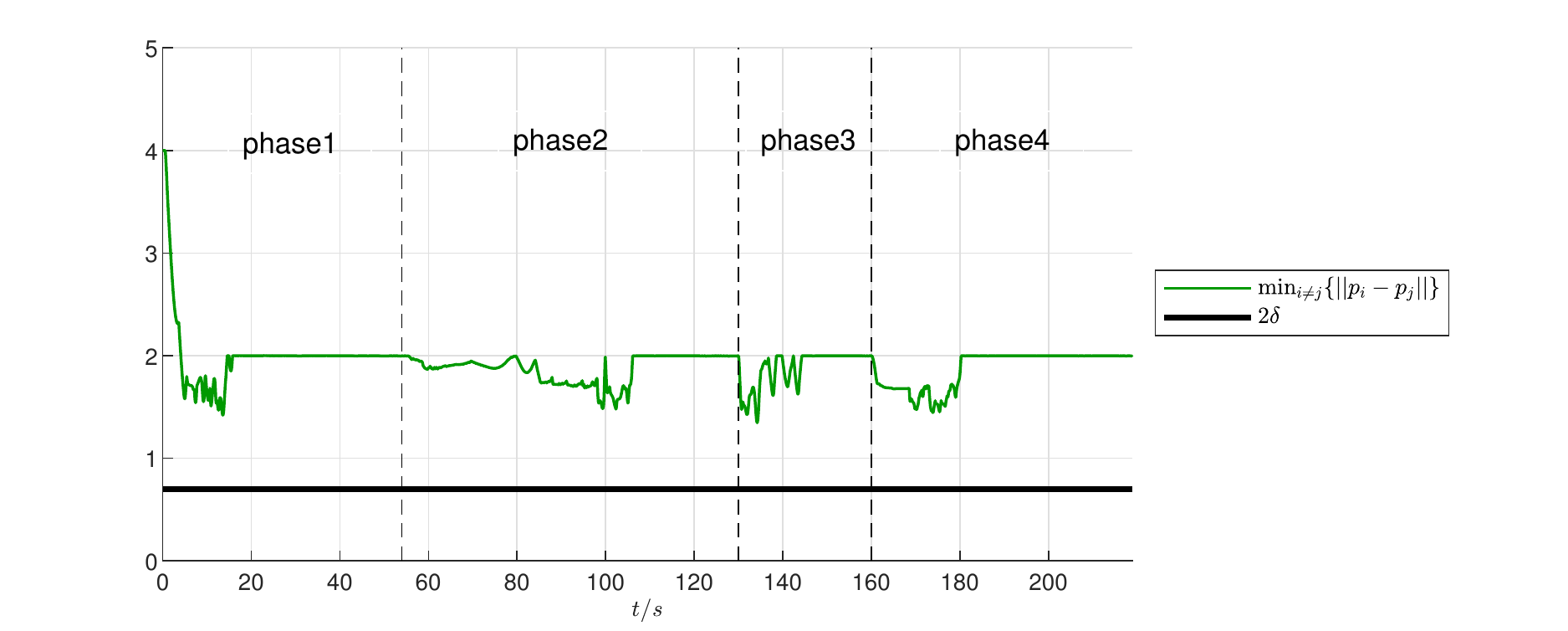}}
    \end{minipage}%
} 
\end{center}
\caption{Relative distance between each robot and the mobile obstacle, and the minimal relative distance between two robots by ROS simulation.}\label{ROS_collision_avoidance}
\end{figure*}

 Figs. \ref{ROS_phase1}-\ref{ROS_phase4} show the positions of the robot swarm and the obstacle
 during each phase, respectively.
 Similar to the simulation results in Section \ref{sec.apptodis}, it can be observed that both
 the line marching objective and the collision avoidance objective have been achieved. Moreover,
 the algorithm proves to be robust against robot failure.
 Figs. \ref{ROS_xyp}  and \ref{ROS_xyv} show the time profiles of
 robots' positions and velocities, respectively.
 Fig. \ref{ROS_collision_avoidance} shows the relative distance between each robot and the mobile obstacle,
 and the minimal relative distance between two robots.
 These simulation results further confirm the
 effectiveness of the proposed algorithm.

\section{Conclusion}

This paper proposed a dynamic leader-follower approach to solving
the line marching problem for a swarm of planar kinematic robots.
In contrast to the existing exact formation algorithms, such as the virtual structure approach
or the traditional leader-follower approach, the proposed approach shows strong robustness
against robot failure by constantly updating the chain of leader-follower pairs.
Comprehensive numerical results using Matlab, Python and ROS
are provided to validate the effectiveness of the
proposed algorithm for continuous-time, discrete-time and unicycle systems, respectively.

\section*{Funding}
This work was supported in part by National Natural Science
Foundation of China under Grant [grant number 61773170, 61703167], and in part by Guangdong Nature Science Foundation under
[grant number 2021A1515012584].

\end{document}